\documentclass[10pt]{article} 
\usepackage{amsthm}
\usepackage{amsmath,amssymb}         
\usepackage{stmaryrd}    
\usepackage{float}

\usepackage{xspace} 
\usepackage{color} 
\usepackage{epsfig}    
\graphicspath{{.}{./figures/}} 

\usepackage{macros/tikzfig}
\usepackage{keycommand} 

\usepackage[ruled,vlined]{algorithm2e}
\usepackage{endnotes}

\newkeycommand{\boxmap}[style=small box][1]{\,\tikz{\node[style=\commandkey{style}] (x) {$#1$};\draw(0,-1.25)--(x)--(0,1.25);}\,}
\newkeycommand{\dboxmap}[style=dbox][1]{\,\tikz{\node[style=\commandkey{style}] (x) {$#1$};\draw[boldedge] (0,-1.25)--(x)--(0,1.25);}\,}

\newkeycommand{\wirelabel}[style=white label]{\,\tikz{\node[style=\commandkey{style}]{};}\,\xspace}

\newkeycommand{\pointmap}[style=point][1]{\,\tikz{\node[style=\commandkey{style}] (x) at (0,-0.3) {$#1$};\node [style=none] (3) at (0, 0.9) {};\node [style=none] (3) at (0, -0.9) {};\draw (x)--(0,0.7);}\,}

\newkeycommand{\point}[style=point,estyle=][1]{\,\tikz{\node[style=\commandkey{style}] (x) at (0,-0.05) {$#1$};\draw [\commandkey{estyle}] (x)--(0,1.05);}\,}

\newkeycommand{\copointmap}[style=copoint][1]{\,\tikz{\node[style=\commandkey{style}] (x) at (0,0.3) {$#1$};\draw (x)--(0,-0.7);}\,}

\newkeycommand{\dpoint}[style=point][1]{\,\tikz{\node[style=\commandkey{style},doubled] (x) at (0,-0.3) {$#1$};\draw[boldedge] (x)--(0,0.7);}\,}

\newkeycommand{\kpoint}[style=kpoint,estyle=][1]{\,\tikz{\node[style=\commandkey{style}] (x) at (0,-0.05) {$#1$};\draw [\commandkey{estyle}] (x)--(0,1.05);}\,}

\newkeycommand{\kpointgray}[style=gray kpoint,estyle=][1]{\,\tikz{\node[style=\commandkey{style}] (x) at (0,-0.05) {$#1$};\draw [\commandkey{estyle}] (x)--(0,1.05);}\,}

\newkeycommand{\typedkpoint}[style=kpoint,edgestyle=][2]{%
\,\begin{tikzpicture}
  \begin{pgfonlayer}{nodelayer}
    \node [style=none] (0) at (0, 1) {};
    \node [style=\commandkey{style}] (1) at (0, -0.75) {$#2$};
    \node [style=right label] (2) at (0.25, 0.5) {$#1$};
  \end{pgfonlayer}
  \begin{pgfonlayer}{edgelayer}
    \draw [\commandkey{edgestyle}] (1) to (0.center);
  \end{pgfonlayer}
\end{tikzpicture}\,}

\newkeycommand{\typedkpointdag}[style=kpoint adjoint,edgestyle=][2]{%
\,\begin{tikzpicture}
  \begin{pgfonlayer}{nodelayer}
    \node [style=none] (0) at (0, -1) {};
    \node [style=\commandkey{style}] (1) at (0, 0.75) {$#2$};
    \node [style=right label] (2) at (0.25, -0.5) {$#1$};
  \end{pgfonlayer}
  \begin{pgfonlayer}{edgelayer}
    \draw [\commandkey{edgestyle}] (0.center) to (1);
  \end{pgfonlayer}
\end{tikzpicture}\,}

\newkeycommand{\bistate}[style=kpoint][1]{\,\begin{tikzpicture}
  \begin{pgfonlayer}{nodelayer}
    \node [style=\commandkey{style}, minimum width=1 cm, inner sep=2pt] (0) at (0, -0.25) {$#1$};
    \node [style=none] (1) at (-0.75, 0) {};
    \node [style=none] (2) at (0.75, 0) {};
    \node [style=none] (3) at (-0.75, 0.88) {};
    \node [style=none] (4) at (0.75, 0.88) {};
  \end{pgfonlayer}
  \begin{pgfonlayer}{edgelayer}
    \draw (1.center) to (3.center);
    \draw (2.center) to (4.center);
  \end{pgfonlayer}
\end{tikzpicture}\,}

\newkeycommand{\bistatebig}[style=kpoint][1]{\,\begin{tikzpicture}
  \begin{pgfonlayer}{nodelayer}
    \node [style=\commandkey{style}, minimum width=, minimum width=1.26 cm, inner sep=2pt] (0) at (0, -0.25) {$#1$};
    \node [style=none] (1) at (-1, 0) {};
    \node [style=none] (2) at (1, 0) {};
    \node [style=none] (3) at (-1, 0.88) {};
    \node [style=none] (4) at (1, 0.88) {};
  \end{pgfonlayer}
  \begin{pgfonlayer}{edgelayer}
    \draw (1.center) to (3.center);
    \draw (2.center) to (4.center);
  \end{pgfonlayer}
\end{tikzpicture}\,}

\newkeycommand{\dbistate}[style=dkpoint][1]{\,\begin{tikzpicture}
  \begin{pgfonlayer}{nodelayer}
    \node [style=\commandkey{style}, minimum width=1 cm, inner sep=2pt] (0) at (0, -0.25) {$#1$};
    \node [style=none] (1) at (-0.75, 0) {};
    \node [style=none] (2) at (0.75, 0) {};
    \node [style=none] (3) at (-0.75, 1.0) {};
    \node [style=none] (4) at (0.75, 1.0) {};
  \end{pgfonlayer}
  \begin{pgfonlayer}{edgelayer}
    \draw [boldedge] (1.center) to (3.center);
    \draw [boldedge] (2.center) to (4.center);
  \end{pgfonlayer}
\end{tikzpicture}\,}

\newkeycommand{\bistatebraket}[style1=kpoint,style2=kpointadj][2]{\,\begin{tikzpicture}
  \begin{pgfonlayer}{nodelayer}
    \node [style=\commandkey{style1}, minimum width=1 cm, inner sep=2pt] (0) at (0, -0.75) {$#1$};
    \node [style=none] (1) at (-0.75, -0.5) {};
    \node [style=none] (2) at (0.75, -0.5) {};
    \node [style=none] (3) at (-0.75, 0.5) {};
    \node [style=none] (4) at (0.75, 0.5) {};
    \node [style=\commandkey{style2}, minimum width=1 cm, inner sep=2pt] (5) at (0, 0.75) {$#2$};
  \end{pgfonlayer}
  \begin{pgfonlayer}{edgelayer}
    \draw (1.center) to (3.center);
    \draw (2.center) to (4.center);
  \end{pgfonlayer}
\end{tikzpicture}\,}

\newkeycommand{\weight}[style=dkpoint][1]{\,\begin{tikzpicture}
  \begin{pgfonlayer}{nodelayer}
    \node [style=\commandkey{style}] (0) at (0, -0.5) {$#1$};
    \node [style=upground] (1) at (0, 0.75) {};
  \end{pgfonlayer}
  \begin{pgfonlayer}{edgelayer}
    \draw [style=boldedge] (0) to (1);
  \end{pgfonlayer}
\end{tikzpicture}\,}

\newkeycommand{\dkpoint}[style=dkpoint][1]{\,\tikz{\node[style=\commandkey{style}] (x) at (0,-0.1) {$#1$};\draw[boldedge] (x)--(0,1);}\,}
\newkeycommand{\dkpointadj}[style=dkpointadj][1]{\,\tikz{\node[style=\commandkey{style}] (x) at (0,0.1) {$#1$};\draw[boldedge] (x)--(0,-1);}\,}
\newkeycommand{\dkpointtrans}[style=dkpointtrans][1]{\,\tikz{\node[style=\commandkey{style}] (x) at (0,0.1) {$#1$};\draw[boldedge] (x)--(0,-1);}\,}
\newkeycommand{\dkpointconj}[style=dkpointconj][1]{\,\tikz{\node[style=\commandkey{style}] (x) at (0,-0.1) {$#1$};\draw[boldedge] (x)--(0,1);}\,}

\newkeycommand{\blackkpoint}[style=black kpoint][1]{\,\tikz{\node[style=\commandkey{style}] (x) at (0,-0.1) {$#1$};\draw (x)--(0,1);}\,}
\newkeycommand{\blackkpointadj}[style=black kpointadj][1]{\,\tikz{\node[style=\commandkey{style}] (x) at (0,0.1) {$#1$};\draw (x)--(0,-1);}\,}
\newkeycommand{\blackdkpoint}[style=black dkpoint][1]{\,\tikz{\node[style=\commandkey{style}] (x) at (0,-0.1) {$#1$};\draw[boldedge] (x)--(0,1);}\,}
\newkeycommand{\blackdkpointadj}[style=black dkpointadj][1]{\,\tikz{\node[style=\commandkey{style}] (x) at (0,0.1) {$#1$};\draw[boldedge] (x)--(0,-1);}\,}

\newcommand{\trace}{\,\begin{tikzpicture}
  \begin{pgfonlayer}{nodelayer}
    \node [style=none] (0) at (0, -0.60) {};
    \node [style=upground] (1) at (0, 0.40) {};
  \end{pgfonlayer}
  \begin{pgfonlayer}{edgelayer}
    \draw [style=none] (0.center) to (1);
  \end{pgfonlayer}
\end{tikzpicture}\,}

\newcommand\discard\trace

\newkeycommand{\pointketbra}[style1=copoint,style2=point,estyle=][2]{\,%
\begin{tikzpicture}
  \begin{pgfonlayer}{nodelayer}
    \node [style=none] (0) at (0, -1.75) {};
    \node [style=\commandkey{style1}] (1) at (0, -0.75) {$#1$};
    \node [style=\commandkey{style2}] (2) at (0, 0.75) {$#2$};
    \node [style=none] (3) at (0, 1.75) {};
  \end{pgfonlayer}
  \begin{pgfonlayer}{edgelayer}
    \draw [\commandkey{estyle}] (0.center) to (1);
    \draw [\commandkey{estyle}] (2) to (3.center);
  \end{pgfonlayer}
\end{tikzpicture}\,}

\newkeycommand{\twocopointketbra}[style1=copoint,style2=copoint,style3=point][3]{\,%
\begin{tikzpicture}
  \begin{pgfonlayer}{nodelayer}
    \node [style=none] (0) at (-0.75, -1.75) {};
    \node [style=none] (1) at (0.75, -1.75) {};
    \node [style=\commandkey{style1}] (2) at (-0.75, -0.75) {$#1$};
    \node [style=\commandkey{style2}] (3) at (0.75, -0.75) {$#2$};
    \node [style=\commandkey{style3}] (4) at (0, 0.75) {$#3$};
    \node [style=none] (5) at (0, 1.75) {};
  \end{pgfonlayer}
  \begin{pgfonlayer}{edgelayer}
    \draw (0.center) to (2);
    \draw (1.center) to (3);
    \draw (4) to (5.center);
  \end{pgfonlayer}
\end{tikzpicture}\,}

\newkeycommand{\twopointketbra}[style1=copoint,style2=point,style3=point][3]{\,%
\begin{tikzpicture}
  \begin{pgfonlayer}{nodelayer}
    \node [style=none] (0) at (0, -1.75) {};
    \node [style=none] (1) at (-0.75, 1.75) {};
    \node [style=\commandkey{style1}] (2) at (0, -0.75) {$#1$};
    \node [style=\commandkey{style2}] (3) at (-0.75, 0.75) {$#2$};
    \node [style=\commandkey{style3}] (4) at (0.75, 0.75) {$#3$};
    \node [style=none] (5) at (0.75, 1.75) {};
  \end{pgfonlayer}
  \begin{pgfonlayer}{edgelayer}
    \draw (0.center) to (2);
    \draw (1.center) to (3);
    \draw (4) to (5.center);
  \end{pgfonlayer}
\end{tikzpicture}\,}

\newkeycommand{\pointbraket}[style1=point,style2=copoint,estyle=][2]{\,%
\begin{tikzpicture}
  \begin{pgfonlayer}{nodelayer}
    \node [style=\commandkey{style1}] (0) at (0, -0.625) {$#1$};
    \node [style=\commandkey{style2}] (1) at (0, 0.625) {$#2$};
  \end{pgfonlayer}
  \begin{pgfonlayer}{edgelayer}
    \draw [\commandkey{estyle}] (0) to (1);
  \end{pgfonlayer}
\end{tikzpicture}\,}

\newkeycommand{\kpointketbra}[style1=kpointdag,style2=kpoint,estyle=][2]{\,%
\begin{tikzpicture}
  \begin{pgfonlayer}{nodelayer}
    \node [style=none] (0) at (0, -1.9) {};
    \node [style=\commandkey{style1}] (1) at (0, -0.95) {$#1$};
    \node [style=\commandkey{style2}] (2) at (0, 0.95) {$#2$};
    \node [style=none] (3) at (0, 1.9) {};
  \end{pgfonlayer}
  \begin{pgfonlayer}{edgelayer}
    \draw [\commandkey{estyle}] (0.center) to (1);
    \draw [\commandkey{estyle}] (2) to (3.center);
  \end{pgfonlayer}
\end{tikzpicture}\,}

\newkeycommand{\kpointmap}[style1=kpoint,style2=map][2]{\,%
\begin{tikzpicture}
  \begin{pgfonlayer}{nodelayer}
    \node [style=\commandkey{style1}] (0) at (0, -1.1) {$#1$};
    \node [style=\commandkey{style2}] (1) at (0, 0.5) {$#2$};
    \node [style=none] (2) at (0, 1.75) {};
  \end{pgfonlayer}
  \begin{pgfonlayer}{edgelayer}
    \draw (0) to (1);
    \draw (1) to (2.center);
  \end{pgfonlayer}
\end{tikzpicture}%
\,}

\newkeycommand{\kpointmaptrans}[style1=kpointconj,style2=mapadj][2]{\,%
\begin{tikzpicture}
  \begin{pgfonlayer}{nodelayer}
    \node [style=\commandkey{style1}] (0) at (0, -1.1) {$#1$};
    \node [style=\commandkey{style2}] (1) at (0, 0.5) {$#2$};
    \node [style=none] (2) at (0, 1.75) {};
  \end{pgfonlayer}
  \begin{pgfonlayer}{edgelayer}
    \draw (0) to (1);
    \draw (1) to (2.center);
  \end{pgfonlayer}
\end{tikzpicture}%
\,}

\newkeycommand{\kpointmapadj}[style1=kpointadj,style2=map][2]{\,%
\begin{tikzpicture}
  \begin{pgfonlayer}{nodelayer}
    \node [style=\commandkey{style1}] (0) at (0, 1.1) {$#1$};
    \node [style=\commandkey{style2}] (1) at (0, -0.5) {$#2$};
    \node [style=none] (2) at (0, -1.75) {};
  \end{pgfonlayer}
  \begin{pgfonlayer}{edgelayer}
    \draw (0) to (1);
    \draw (1) to (2.center);
  \end{pgfonlayer}
\end{tikzpicture}%
\,}

\newkeycommand{\dkpointmap}[style1=dkpoint,style2=dmap][2]{\,%
\begin{tikzpicture}
  \begin{pgfonlayer}{nodelayer}
    \node [style=\commandkey{style1}] (0) at (0, -1.2) {$#1$};
    \node [style=\commandkey{style2}] (1) at (0, 0.4) {$#2$};
    \node [style=none] (2) at (0, 1.7) {};
  \end{pgfonlayer}
  \begin{pgfonlayer}{edgelayer}
    \draw[style=boldedge] (0) to (1);
    \draw[style=boldedge] (1) to (2.center);
  \end{pgfonlayer}
\end{tikzpicture}%
\,}

\newkeycommand{\dkpointmapx}[style1=dkpoint,style2=dmap][2]{\,%
\begin{tikzpicture}
  \begin{pgfonlayer}{nodelayer}
    \node [style=\commandkey{style1}] (0) at (0, -1.4) {$#1$};
    \node [style=\commandkey{style2}] (1) at (0, 0.6) {$#2$};
    \node [style=none] (2) at (0, 1.9) {};
  \end{pgfonlayer}
  \begin{pgfonlayer}{edgelayer}
    \draw[style=boldedge] (0) to (1);
    \draw[style=boldedge] (1) to (2.center);
  \end{pgfonlayer}
\end{tikzpicture}%
\,}

\newkeycommand{\boxcopointmapdag}[style1=map,style2=copoint][2]{\,%
\begin{tikzpicture}
  \begin{pgfonlayer}{nodelayer}
    \node [style=none] (0) at (0, -1.75) {};
    \node [style=\commandkey{style1}] (1) at (0, -0.5) {$#1$};
    \node [style=\commandkey{style2}] (2) at (0, 1) {$#2$};
  \end{pgfonlayer}
  \begin{pgfonlayer}{edgelayer}
    \draw (0.center) to (1);
    \draw (1) to (2);
  \end{pgfonlayer}
\end{tikzpicture}%
\,}

\newkeycommand{\kpointdagmap}[style1=map,style2=kpointdag][2]{\,%
\begin{tikzpicture}
  \begin{pgfonlayer}{nodelayer}
    \node [style=none] (0) at (0, -1.75) {};
    \node [style=\commandkey{style1}] (1) at (0, -0.5) {$#1$};
    \node [style=\commandkey{style2}] (2) at (0, 1.1) {$#2$};
  \end{pgfonlayer}
  \begin{pgfonlayer}{edgelayer}
    \draw (0.center) to (1);
    \draw (1) to (2);
  \end{pgfonlayer}
\end{tikzpicture}%
\,}

\newkeycommand{\sandwichmap}[style1=point,style2=map,style3=copoint][3]{\,%
\begin{tikzpicture}
  \begin{pgfonlayer}{nodelayer}
    \node [style=\commandkey{style1}] (0) at (0, -1.50) {$#1$};
    \node [style=\commandkey{style2}] (1) at (0, 0) {$#2$};
    \node [style=\commandkey{style3}] (2) at (0, 1.50) {$#3$};
  \end{pgfonlayer}
  \begin{pgfonlayer}{edgelayer}
    \draw (0) to (1);
    \draw (1) to (2);
  \end{pgfonlayer}
\end{tikzpicture}%
}

\newkeycommand{\kpointsandwichmap}[style1=kpoint,style2=map,style3=kpointdag][3]{\,%
\begin{tikzpicture}
  \begin{pgfonlayer}{nodelayer}
    \node [style=\commandkey{style1}] (0) at (0, -1.5) {$#1$};
    \node [style=\commandkey{style2}] (1) at (0, 0) {$#2$};
    \node [style=\commandkey{style3}] (2) at (0, 1.5) {$#3$};
  \end{pgfonlayer}
  \begin{pgfonlayer}{edgelayer}
    \draw (0) to (1);
    \draw (1) to (2);
  \end{pgfonlayer}
\end{tikzpicture}%
}

\newkeycommand{\sandwichmapconj}[style1=point,style2=mapconj,style3=copoint][3]{\,%
\begin{tikzpicture}
  \begin{pgfonlayer}{nodelayer}
    \node [style=\commandkey{style1}] (0) at (0, -1.50) {$#1$};
    \node [style=\commandkey{style2}] (1) at (0, 0) {$#2$};
    \node [style=\commandkey{style3}] (2) at (0, 1.50) {$#3$};
  \end{pgfonlayer}
  \begin{pgfonlayer}{edgelayer}
    \draw (0) to (1);
    \draw (1) to (2);
  \end{pgfonlayer}
\end{tikzpicture}%
}

\newkeycommand{\sandwichtwo}[style1=point,style2=map,style3=map,style4=copoint][4]{\,%
\begin{tikzpicture}
  \begin{pgfonlayer}{nodelayer}
    \node [style=\commandkey{style2}] (0) at (0, -1) {$#2$};
    \node [style=\commandkey{style3}] (1) at (0, 1) {$#3$};
    \node [style=\commandkey{style1}] (2) at (0, -2.6) {$#1$};
    \node [style=\commandkey{style4}] (3) at (0, 2.6) {$#4$};
  \end{pgfonlayer}
  \begin{pgfonlayer}{edgelayer}
    \draw (2) to (0);
    \draw (0) to (1);
    \draw (1) to (3);
  \end{pgfonlayer}
\end{tikzpicture}\,}

\newkeycommand{\longbraket}[style1=point,style2=copoint][2]{\,%
\begin{tikzpicture}
  \begin{pgfonlayer}{nodelayer}
    \node [style=\commandkey{style1}] (0) at (0, -2.6) {$#1$};
    \node [style=\commandkey{style2}] (1) at (0, 2.6) {$#2$};
  \end{pgfonlayer}
  \begin{pgfonlayer}{edgelayer}
    \draw (0) to (1);
  \end{pgfonlayer}
\end{tikzpicture}\,}

\newkeycommand{\onb}[style=point]{\ensuremath{\{\,\tikz{\node[style=\commandkey{style}] (x) at (0,-0.3) {$j$};\draw (x)--(0,0.7);}\,\}}\xspace}

\newkeycommand{\phase}[style=white phase dot][1]{\,\begin{tikzpicture}
    \begin{pgfonlayer}{nodelayer}
        \node [style=none] (0) at (0, 1) {};
        \node [style=\commandkey{style}] (2) at (0, -0) {$#1$}; 
        \node [style=none] (3) at (0, -1) {};
    \end{pgfonlayer}
    \begin{pgfonlayer}{edgelayer}
        \draw (2) to (0.center);
        \draw (3.center) to (2);
    \end{pgfonlayer}
\end{tikzpicture}\,}

\newkeycommand{\dphase}[style=white phase ddot][1]{\,\begin{tikzpicture}
    \begin{pgfonlayer}{nodelayer}
        \node [style=none] (0) at (0, 1.25) {};
        \node [style=\commandkey{style}] (2) at (0, -0) {$#1$};
        \node [style=none] (3) at (0, -1.25) {};
    \end{pgfonlayer}
    \begin{pgfonlayer}{edgelayer}
        \draw [boldedge] (2) to (0.center);
        \draw [boldedge] (3.center) to (2);
    \end{pgfonlayer}
\end{tikzpicture}\,}

\newkeycommand{\dphasepoint}[style=white phase ddot][1]{\,\begin{tikzpicture}[yshift=-3mm]
  \begin{pgfonlayer}{nodelayer}
    \node [style=none] (0) at (0, 1) {};
    \node [style=\commandkey{style}] (1) at (0, 0) {$#1$};
  \end{pgfonlayer}
  \begin{pgfonlayer}{edgelayer}
    \draw [style=boldedge] (0.center) to (1);
  \end{pgfonlayer}
\end{tikzpicture}\,}

\newkeycommand{\dphasepointgray}[style=gray phase ddot][1]{\,\begin{tikzpicture}[yshift=-3mm]
  \begin{pgfonlayer}{nodelayer}
    \node [style=none] (0) at (0, 1) {};
    \node [style=\commandkey{style}] (1) at (0, 0) {$#1$};
  \end{pgfonlayer}
  \begin{pgfonlayer}{edgelayer}
    \draw [style=boldedge] (0.center) to (1);
  \end{pgfonlayer}
\end{tikzpicture}\,}

\newkeycommand{\dphasecopoint}[style=white phase ddot][1]{\,\begin{tikzpicture}[yshift=3mm,yscale=-1]
  \begin{pgfonlayer}{nodelayer}
    \node [style=none] (0) at (0, 1) {};
    \node [style=\commandkey{style}] (1) at (0, 0) {$#1$};
  \end{pgfonlayer}
  \begin{pgfonlayer}{edgelayer}
    \draw [style=boldedge] (0.center) to (1);
  \end{pgfonlayer}
\end{tikzpicture}\,}

\newkeycommand{\dphasepointsm}[style=white phase ddot][1]{\,\begin{tikzpicture}[yshift=-3mm]
  \begin{pgfonlayer}{nodelayer}
    \node [style=none] (0) at (0, 1) {};
    \node [style=\commandkey{style}] (1) at (0, 0) {\footnotesize\!$#1$\!};
  \end{pgfonlayer}
  \begin{pgfonlayer}{edgelayer}
    \draw [style=boldedge] (0.center) to (1);
  \end{pgfonlayer}
\end{tikzpicture}\,}

\newkeycommand{\phasepoint}[style=white phase dot][1]{\,\begin{tikzpicture}[yshift=-3mm]
  \begin{pgfonlayer}{nodelayer}
    \node [style=none] (0) at (0, 1) {};
    \node [style=\commandkey{style}] (1) at (0, 0) {$#1$};
  \end{pgfonlayer}
  \begin{pgfonlayer}{edgelayer}
    \draw (0.center) to (1);
  \end{pgfonlayer}
\end{tikzpicture}\,}

\newkeycommand{\phasecopoint}[style=white phase dot][1]{\,\begin{tikzpicture}[yshift=3mm]
  \begin{pgfonlayer}{nodelayer}
    \node [style=none] (0) at (0, -1) {};
    \node [style=\commandkey{style}] (1) at (0, 0) {$#1$};
  \end{pgfonlayer}
  \begin{pgfonlayer}{edgelayer}
    \draw (0.center) to (1);
  \end{pgfonlayer}
\end{tikzpicture}\,}

\newkeycommand{\kpointbraket}[style1=kpoint,style2=kpoint adjoint,estyle=][2]{\,%
\begin{tikzpicture}
  \begin{pgfonlayer}{nodelayer}
    \node [style=\commandkey{style1}] (0) at (0, -0.735) {$#1$};
    \node [style=\commandkey{style2}] (1) at (0, 0.735) {$#2$};
  \end{pgfonlayer}
  \begin{pgfonlayer}{edgelayer}
    \draw [\commandkey{estyle}] (0) to (1);
  \end{pgfonlayer}
\end{tikzpicture}\,}

\newkeycommand{\kkpointbraket}[style1=kpoint,style2=kpoint adjoint,estyle=][2]{\,%
\begin{tikzpicture}
  \begin{pgfonlayer}{nodelayer}
    \node [style=\commandkey{style1}] (0) at (0, -0.685) {$#1$};
    \node [style=\commandkey{style2}] (1) at (0, 0.685) {$#2$};
  \end{pgfonlayer}
  \begin{pgfonlayer}{edgelayer}
    \draw [\commandkey{estyle}] (0) to (1);
  \end{pgfonlayer}
\end{tikzpicture}\,}

\newkeycommand{\kpointbraketx}[style1=kpoint,style2=kpoint adjoint,estyle=][2]{\,%
\begin{tikzpicture}
  \begin{pgfonlayer}{nodelayer}
    \node [style=\commandkey{style1}] (0) at (0, -0.885) {$#1$};
    \node [style=\commandkey{style2}] (1) at (0, 0.885) {$#2$};
  \end{pgfonlayer}
  \begin{pgfonlayer}{edgelayer}
    \draw [\commandkey{estyle}] (0) to (1);
  \end{pgfonlayer}
\end{tikzpicture}\,}

\tikzstyle{cdiag}=[matrix of math nodes, row sep=3em, column sep=3em, text height=1.5ex, text depth=0.25ex,inner sep=0.5em]
\tikzstyle{arrow above}=[transform canvas={yshift=0.5ex}]
\tikzstyle{arrow below}=[transform canvas={yshift=-0.5ex}]

\usepackage{tikz}
\usetikzlibrary{decorations.markings}
\usetikzlibrary{shapes.geometric}

\pgfdeclarelayer{edgelayer}
\pgfdeclarelayer{nodelayer}
\pgfsetlayers{edgelayer,nodelayer,main}

\tikzstyle{none}=[inner sep=0pt]
\definecolor{hexcolor0xff0000}{rgb}{1.000,0.000,0.000}
\definecolor{hexcolor0x000000}{rgb}{0.000,0.000,0.000}
\definecolor{hexcolor0x00ff00}{rgb}{0.000,1.000,0.000}
\definecolor{hexcolor0x000000}{rgb}{0.000,0.000,0.000}
\definecolor{hexcolor0xffff00}{rgb}{1.000,1.000,0.000}
\definecolor{hexcolor0xffffff}{rgb}{1.000,1.000,1.000}

\tikzstyle{rn}=[circle,fill=hexcolor0xff0000,draw=hexcolor0x000000,line width=0.8 pt]
\tikzstyle{gn}=[circle,fill=hexcolor0x00ff00,draw=hexcolor0x000000,line width=0.8 pt]
\tikzstyle{yn}=[circle,fill=hexcolor0xffff00,draw=hexcolor0x000000,line width=0.8 pt]
\tikzstyle{wn}=[circle,fill=hexcolor0xffffff,draw=hexcolor0x000000,line width=0.8 pt]
\tikzstyle{wnthick}=[circle,fill=hexcolor0xffffff,draw=hexcolor0x000000,line width=2.500]

\tikzstyle{simple}=[-,draw=hexcolor0x000000,line width=2.000]
\tikzstyle{arrow}=[-,draw=hexcolor0x000000,postaction={decorate},decoration={markings,mark=at position .5 with {\arrow{>}}},line width=2.000]
\tikzstyle{tick}=[-,draw=hexcolor0x000000,postaction={decorate},decoration={markings,mark=at position .5 with {\draw (0,-0.1) -- (0,0.1);}},line width=2.000]
\tikzstyle{halfthickness}=[-,draw=hexcolor0x000000,line width=0.500]
\tikzstyle{thick}=[-,draw=hexcolor0x000000,line width=2.500]
\tikzstyle{thicker}=[-,draw=hexcolor0x000000,line width=4.000]

\tikzstyle{env}=[copoint,regular polygon rotate=0,minimum width=0.2cm, fill=black]

\tikzstyle{probs}=[shape=semicircle,fill=white,draw=black,shape border rotate=180,minimum width=1.2cm]

\tikzstyle{every picture}=[baseline=-0.25em,scale=0.5]
\tikzstyle{dotpic}=[] 
\tikzstyle{diredges}=[every to/.style={diredge}]
\tikzstyle{math matrix}=[matrix of math nodes,left delimiter=(,right delimiter=),inner sep=2pt,column sep=1em,row sep=0.5em,nodes={inner sep=0pt},text height=1.5ex, text depth=0.25ex]

\tikzstyle{inline text}=[text height=1.5ex, text depth=0.25ex,yshift=0.5mm]
\tikzstyle{label}=[font=\footnotesize,text height=1.5ex, text depth=0.25ex,yshift=0.5mm]
\tikzstyle{left label}=[label,anchor=east,xshift=1.5mm]
\tikzstyle{right label}=[label,anchor=west,xshift=-1.5mm]

\tikzstyle{braceedge}=[decorate,decoration={brace,amplitude=2mm,raise=-1mm}]
\tikzstyle{small braceedge}=[decorate,decoration={brace,amplitude=1mm,raise=-1mm}]

\tikzstyle{doubled}=[line width=1.6pt] 
\tikzstyle{boldedge}=[doubled,shorten <=-0.17mm,shorten >=-0.17mm]
\tikzstyle{boldedgegray}=[doubled,gray,shorten <=-0.17mm,shorten >=-0.17mm]

\tikzstyle{semidoubled}=[line width=1.4pt] 
\tikzstyle{semiboldedgegray}=[semidoubled,gray,shorten <=-0.17mm,shorten >=-0.17mm]

\tikzstyle{boldedgedashed}=[very thick,dashed,shorten <=-0.17mm,shorten >=-0.17mm]
\tikzstyle{vboldedgedashed}=[doubled,dashed,shorten <=-0.17mm,shorten >=-0.17mm]
\tikzstyle{left hook arrow}=[left hook-latex]
\tikzstyle{right hook arrow}=[right hook-latex]
\tikzstyle{sembracket}=[line width=0.5pt,shorten <=-0.07mm,shorten >=-0.07mm]

\tikzstyle{causal edge}=[->,thick,gray]
\tikzstyle{causal nondir}=[thick,gray]
\tikzstyle{timeline}=[thick,gray, dashed]

\tikzstyle{cedge}=[<->,thick,gray!70!white]

\tikzstyle{empty diagram}=[draw=gray!40!white,dashed,shape=rectangle,minimum width=1cm,minimum height=1cm]
\tikzstyle{empty diagram small}=[draw=gray!50!white,dashed,shape=rectangle,minimum width=0.6cm,minimum height=0.5cm]

\tikzstyle{dot}=[inner sep=0mm,minimum width=2mm,minimum height=2mm,draw,shape=circle]
\tikzstyle{ddot}=[inner sep=0mm, doubled, minimum width=2.5mm,minimum height=2.5mm,draw,shape=circle]

\tikzstyle{black dot}=[dot,fill=black]
\tikzstyle{white dot}=[dot,fill=white,,text depth=-0.2mm]
\tikzstyle{green dot}=[white dot] 
\tikzstyle{gray dot}=[dot,fill=gray!40!white,,text depth=-0.2mm]
\tikzstyle{red dot}=[gray dot] 

\tikzstyle{black ddot}=[ddot,fill=black]
\tikzstyle{white ddot}=[ddot,fill=white]
\tikzstyle{gray ddot}=[ddot,fill=gray!40!white]

\tikzstyle{gray edge}=[gray!40!white]

\tikzstyle{small dot}=[inner sep=0.5mm,minimum width=0pt,minimum height=0pt,draw,shape=circle]

\tikzstyle{small black dot}=[small dot,fill=black]
\tikzstyle{small white dot}=[small dot,fill=white]
\tikzstyle{small gray dot}=[small dot,fill=gray!40!white]

\tikzstyle{causal dot}=[inner sep=0.4mm,minimum width=0pt,minimum height=0pt,draw=white,shape=circle,fill=gray!40!white]

\tikzstyle{phase dimensions}=[minimum size=5mm,font=\footnotesize,rectangle,rounded corners=2.5mm,inner sep=0.2mm,outer sep=-2mm]
\tikzstyle{dphase dimensions}=[minimum size=5mm,font=\footnotesize,rectangle,rounded corners=2.5mm,inner sep=0.2mm,outer sep=-2mm]

\tikzstyle{white phase dot}=[dot,fill=white,phase dimensions]
\tikzstyle{white phase ddot}=[ddot,fill=white,dphase dimensions]
\tikzstyle{green phase ddot}=[ddot,fill=green,dphase dimensions]

\tikzstyle{white rect ddot}=[draw=black,fill=white,doubled,minimum size=5mm,font=\footnotesize,rectangle,rounded corners=2.5mm,inner sep=0.2mm]
\tikzstyle{gray rect ddot}=[draw=black,fill=gray!40!white,doubled,minimum size=6mm,font=\footnotesize,rectangle,rounded corners=3mm]

\tikzstyle{gray phase dot}=[dot,fill=gray!40!white,phase dimensions]
\tikzstyle{gray phase ddot}=[ddot,fill=gray!40!white,dphase dimensions]
\tikzstyle{red phase ddot}=[ddot,fill=red,dphase dimensions]

\tikzstyle{grey phase dot}=[gray phase dot]
\tikzstyle{grey phase ddot}=[gray phase ddot]

\tikzstyle{small phase dimensions}=[minimum size=4mm,font=\tiny,rectangle,rounded corners=2mm,inner sep=0.2mm,outer sep=-2mm]
\tikzstyle{small dphase dimensions}=[minimum size=4mm,font=\tiny,rectangle,rounded corners=2mm,inner sep=0.2mm,outer sep=-2mm]

\tikzstyle{small gray phase dot}=[dot,fill=gray!40!white,small phase dimensions]
\tikzstyle{small gray phase ddot}=[ddot,fill=gray!40!white,small dphase dimensions]

\tikzstyle{small map}=[draw,shape=rectangle,minimum height=4mm,minimum width=4mm,fill=white]

\tikzstyle{cnot}=[fill=white,shape=circle,inner sep=-1.4pt]

\tikzstyle{asym hadamard}=[fill=white,draw,shape=NEbox,inner sep=0.6mm,font=\footnotesize,minimum height=4mm]
\tikzstyle{asym hadamard conj}=[fill=white,draw,shape=NWbox,inner sep=0.6mm,font=\footnotesize,minimum height=4mm]
\tikzstyle{asym hadamard dag}=[fill=white,draw,shape=SEbox,inner sep=0.6mm,font=\footnotesize,minimum height=4mm]

\tikzstyle{hadamard}=[fill=white,draw,inner sep=0.6mm,font=\footnotesize,minimum height=4mm,minimum width=4mm]
\tikzstyle{small hadamard}=[fill=white,draw,inner sep=0.6mm,minimum height=1.5mm,minimum width=1.5mm]
\tikzstyle{dhadamard}=[hadamard,doubled]
\tikzstyle{small dhadamard}=[small hadamard,doubled]
\tikzstyle{small dhadamard rotate}=[small hadamard,doubled,rotate=45]
\tikzstyle{antipode}=[white dot,inner sep=0.3mm,font=\footnotesize]

\tikzstyle{scalar}=[diamond,draw,inner sep=0.5pt,font=\small]
\tikzstyle{dscalar}=[diamond,doubled, draw,inner sep=0.5pt,font=\small]

\tikzstyle{small box}=[rectangle,inline text,fill=white,draw,minimum height=5mm,yshift=-0.5mm,minimum width=5mm,font=\small]
\tikzstyle{small gray box}=[small box,fill=gray!30]
\tikzstyle{medium box}=[rectangle,inline text,fill=white,draw,minimum height=5mm,yshift=-0.5mm,minimum width=10mm,font=\small]
\tikzstyle{square box}=[small box] 
\tikzstyle{medium gray box}=[small box,fill=gray!30]
\tikzstyle{semilarge box}=[rectangle,inline text,fill=white,draw,minimum height=5mm,yshift=-0.5mm,minimum width=12.5mm,font=\small]
\tikzstyle{large box}=[rectangle,inline text,fill=white,draw,minimum height=5mm,yshift=-0.5mm,minimum width=15mm,font=\small]
\tikzstyle{large gray box}=[small box,fill=gray!30]

\tikzstyle{Bayes box}=[rectangle,fill=black,draw, minimum height=3mm, minimum width=3mm]

\tikzstyle{gray square point}=[small box,fill=gray!50]

\tikzstyle{dphase box white}=[dhadamard]
\tikzstyle{dphase box gray}=[dhadamard,fill=gray!50!white]

\tikzstyle{point}=[regular polygon,regular polygon sides=3,draw,scale=0.75,inner sep=-0.5pt,minimum width=9mm,fill=white,regular polygon rotate=180]
\tikzstyle{copoint}=[regular polygon,regular polygon sides=3,draw,scale=0.75,inner sep=-0.5pt,minimum width=9mm,fill=white]
\tikzstyle{dpoint}=[point,doubled]
\tikzstyle{dcopoint}=[copoint,doubled]

\tikzstyle{wide copoint}=[fill=white,draw,shape=isosceles triangle,shape border rotate=90,isosceles triangle stretches=true,inner sep=0pt,minimum width=1.5cm,minimum height=6.12mm]
\tikzstyle{wide point}=[fill=white,draw,shape=isosceles triangle,shape border rotate=-90,isosceles triangle stretches=true,inner sep=0pt,minimum width=1.5cm,minimum height=6.12mm,yshift=-0.0mm]
\tikzstyle{wide point plus}=[fill=white,draw,shape=isosceles triangle,shape border rotate=-90,isosceles triangle stretches=true,inner sep=0pt,minimum width=1.74cm,minimum height=7mm,yshift=-0.0mm]

\tikzstyle{wide dpoint}=[fill=white,doubled,draw,shape=isosceles triangle,shape border rotate=-90,isosceles triangle stretches=true,inner sep=0pt,minimum width=1.5cm,minimum height=6.12mm,yshift=-0.0mm]
\tikzstyle{wide dcopoint}=[fill=white,doubled,draw,shape=isosceles triangle,shape border rotate=90,isosceles triangle stretches=true,inner sep=0pt,minimum width=1.5cm,minimum height=6.12mm,yshift=-0.0mm]

\tikzstyle{tinypoint}=[regular polygon,regular polygon sides=3,draw,scale=0.55,inner sep=-0.15pt,minimum width=6mm,fill=white,regular polygon rotate=180]

\tikzstyle{white point}=[point]
\tikzstyle{white dpoint}=[dpoint]
\tikzstyle{green point}=[white point] 
\tikzstyle{white copoint}=[copoint]
\tikzstyle{gray point}=[point,fill=gray!40!white]
\tikzstyle{gray dpoint}=[gray point,doubled]
\tikzstyle{red point}=[gray point] 
\tikzstyle{gray copoint}=[copoint,fill=gray!40!white]
\tikzstyle{gray dcopoint}=[gray copoint,doubled]

\tikzstyle{white point guide}=[regular polygon,regular polygon sides=3,font=\scriptsize,draw,scale=0.65,inner sep=-0.5pt,minimum width=9mm,fill=white,regular polygon rotate=180]

\tikzstyle{black point}=[point,fill=black,font=\color{white}]
\tikzstyle{black copoint}=[copoint,fill=black,font=\color{white}]

\tikzstyle{tiny gray point}=[tinypoint,fill=gray!40!white]

\tikzstyle{diredge}=[->]
\tikzstyle{ddiredge}=[<->]
\tikzstyle{rdiredge}=[<-]
\tikzstyle{thickdiredge}=[->, very thick]
\tikzstyle{pointer edge}=[->,very thick,gray]
\tikzstyle{pointer edge part}=[very thick,gray]
\tikzstyle{dashed edge}=[dashed]
\tikzstyle{thick dashed edge}=[very thick,dashed]
\tikzstyle{thick gray dashed edge}=[thick dashed edge,gray!40]
\tikzstyle{thick map edge}=[very thick,|->]

\makeatletter
\newcommand{\boxshape}[3]{%
\pgfdeclareshape{#1}{
\inheritsavedanchors[from=rectangle] 
\inheritanchorborder[from=rectangle]
\inheritanchor[from=rectangle]{center}
\inheritanchor[from=rectangle]{north}
\inheritanchor[from=rectangle]{south}
\inheritanchor[from=rectangle]{west}
\inheritanchor[from=rectangle]{east}
\backgroundpath{
\southwest \pgf@xa=\pgf@x \pgf@ya=\pgf@y
\northeast \pgf@xb=\pgf@x \pgf@yb=\pgf@y

\@tempdima=#2
\@tempdimb=#3

\pgfpathmoveto{\pgfpoint{\pgf@xa - 5pt + \@tempdima}{\pgf@ya}}
\pgfpathlineto{\pgfpoint{\pgf@xa - 5pt - \@tempdima}{\pgf@yb}}
\pgfpathlineto{\pgfpoint{\pgf@xb + 5pt + \@tempdimb}{\pgf@yb}}
\pgfpathlineto{\pgfpoint{\pgf@xb + 5pt - \@tempdimb}{\pgf@ya}}
\pgfpathlineto{\pgfpoint{\pgf@xa - 5pt + \@tempdima}{\pgf@ya}}
\pgfpathclose
}
}}

\boxshape{NEbox}{0pt}{5pt}
\boxshape{SEbox}{0pt}{-5pt}
\boxshape{NWbox}{5pt}{0pt}
\boxshape{SWbox}{-5pt}{0pt}
\boxshape{EBox}{-3pt}{3pt}
\boxshape{WBox}{3pt}{-3pt}
\makeatother

\tikzstyle{cloud}=[shape=cloud,draw,minimum width=1.5cm,minimum height=1.5cm]

\tikzstyle{map}=[draw,shape=NEbox,inner sep=2pt,minimum height=6mm,fill=white]
\tikzstyle{dashedmap}=[draw,dashed,shape=NEbox,inner sep=2pt,minimum height=6mm,fill=white]
\tikzstyle{mapdag}=[draw,shape=SEbox,inner sep=2pt,minimum height=6mm,fill=white]
\tikzstyle{mapadj}=[draw,shape=SEbox,inner sep=2pt,minimum height=6mm,fill=white]
\tikzstyle{maptrans}=[draw,shape=SWbox,inner sep=2pt,minimum height=6mm,fill=white]
\tikzstyle{mapconj}=[draw,shape=NWbox,inner sep=2pt,minimum height=6mm,fill=white]

\tikzstyle{langmap}=[draw,shape=NEbox,inner sep=2pt,minimum height=2.4mm,minimum width=3.2mm,fill=white]
\tikzstyle{langmaptrans}=[draw,shape=SWbox,inner sep=2pt,minimum height=2.4mm,minimum width=3.2mm,fill=white]

\tikzstyle{medium map}=[draw,shape=NEbox,inner sep=2pt,minimum height=6mm,fill=white,minimum width=7mm]
\tikzstyle{medium map dag}=[draw,shape=SEbox,inner sep=2pt,minimum height=6mm,fill=white,minimum width=7mm]
\tikzstyle{medium map adj}=[draw,shape=SEbox,inner sep=2pt,minimum height=6mm,fill=white,minimum width=7mm]
\tikzstyle{medium map trans}=[draw,shape=SWbox,inner sep=2pt,minimum height=6mm,fill=white,minimum width=7mm]
\tikzstyle{medium map conj}=[draw,shape=NWbox,inner sep=2pt,minimum height=6mm,fill=white,minimum width=7mm]
\tikzstyle{semilarge map}=[draw,shape=NEbox,inner sep=2pt,minimum height=6mm,fill=white,minimum width=9.5mm]
\tikzstyle{semilarge map trans}=[draw,shape=SWbox,inner sep=2pt,minimum height=6mm,fill=white,minimum width=9.5mm]
\tikzstyle{semilarge map adj}=[draw,shape=SEbox,inner sep=2pt,minimum height=6mm,fill=white,minimum width=9.5mm]
\tikzstyle{semilarge map dag}=[draw,shape=SEbox,inner sep=2pt,minimum height=6mm,fill=white,minimum width=9.5mm]
\tikzstyle{semilarge map conj}=[draw,shape=NWbox,inner sep=2pt,minimum height=6mm,fill=white,minimum width=9.5mm]
\tikzstyle{large map}=[draw,shape=NEbox,inner sep=2pt,minimum height=6mm,fill=white,minimum width=12mm]
\tikzstyle{large map conj}=[draw,shape=NWbox,inner sep=2pt,minimum height=6mm,fill=white,minimum width=12mm]
\tikzstyle{very large map}=[draw,shape=NEbox,inner sep=2pt,minimum height=6mm,fill=white,minimum width=17mm]

\tikzstyle{medium dmap}=[draw,doubled,shape=NEbox,inner sep=2pt,minimum height=6mm,fill=white,minimum width=7mm]
\tikzstyle{medium dmap dag}=[draw,doubled,shape=SEbox,inner sep=2pt,minimum height=6mm,fill=white,minimum width=7mm]
\tikzstyle{medium dmap adj}=[draw,doubled,shape=SEbox,inner sep=2pt,minimum height=6mm,fill=white,minimum width=7mm]
\tikzstyle{medium dmap trans}=[draw,doubled,shape=SWbox,inner sep=2pt,minimum height=6mm,fill=white,minimum width=7mm]
\tikzstyle{medium dmap conj}=[draw,doubled,shape=NWbox,inner sep=2pt,minimum height=6mm,fill=white,minimum width=7mm]
\tikzstyle{semilarge dmap}=[draw,doubled,shape=NEbox,inner sep=2pt,minimum height=6mm,fill=white,minimum width=9.5mm]
\tikzstyle{semilarge dmap trans}=[draw,doubled,shape=SWbox,inner sep=2pt,minimum height=6mm,fill=white,minimum width=9.5mm]
\tikzstyle{semilarge dmap adj}=[draw,doubled,shape=SEbox,inner sep=2pt,minimum height=6mm,fill=white,minimum width=9.5mm]
\tikzstyle{semilarge dmap dag}=[draw,doubled,shape=SEbox,inner sep=2pt,minimum height=6mm,fill=white,minimum width=9.5mm]
\tikzstyle{semilarge dmap conj}=[draw,doubled,shape=NWbox,inner sep=2pt,minimum height=6mm,fill=white,minimum width=9.5mm]
\tikzstyle{large dmap}=[draw,doubled,shape=NEbox,inner sep=2pt,minimum height=6mm,fill=white,minimum width=12mm]
\tikzstyle{large dmap conj}=[draw,doubled,shape=NWbox,inner sep=2pt,minimum height=6mm,fill=white,minimum width=12mm]
\tikzstyle{large dmap trans}=[draw,doubled,shape=SWbox,inner sep=2pt,minimum height=6mm,fill=white,minimum width=12mm]
\tikzstyle{large dmap adj}=[draw,doubled,shape=SEbox,inner sep=2pt,minimum height=6mm,fill=white,minimum width=12mm]
\tikzstyle{large dmap dag}=[draw,doubled,shape=SEbox,inner sep=2pt,minimum height=6mm,fill=white,minimum width=12mm]
\tikzstyle{very large dmap}=[draw,doubled,shape=NEbox,inner sep=2pt,minimum height=6mm,fill=white,minimum width=19.5mm]

\tikzstyle{muxbox}=[draw,shape=rectangle,minimum height=3mm,minimum width=3mm,fill=white]
\tikzstyle{dmuxbox}=[muxbox,doubled]

\tikzstyle{box}=[draw,shape=rectangle,inner sep=2pt,minimum height=6mm,minimum width=6mm,fill=white]
\tikzstyle{dbox}=[draw,doubled,shape=rectangle,inner sep=2pt,minimum height=6mm,minimum width=6mm,fill=white]
\tikzstyle{dmap}=[draw,doubled,shape=NEbox,inner sep=2pt,minimum height=6mm,fill=white]
\tikzstyle{dmapdag}=[draw,doubled,shape=SEbox,inner sep=2pt,minimum height=6mm,fill=white]
\tikzstyle{dmapadj}=[draw,doubled,shape=SEbox,inner sep=2pt,minimum height=6mm,fill=white]
\tikzstyle{dmaptrans}=[draw,doubled,shape=SWbox,inner sep=2pt,minimum height=6mm,fill=white]
\tikzstyle{dmapconj}=[draw,doubled,shape=NWbox,inner sep=2pt,minimum height=6mm,fill=white]

\tikzstyle{ddmap}=[draw,doubled,dashed,shape=NEbox,inner sep=2pt,minimum height=6mm,fill=white]
\tikzstyle{ddmapdag}=[draw,doubled,dashed,shape=SEbox,inner sep=2pt,minimum height=6mm,fill=white]
\tikzstyle{ddmapadj}=[draw,doubled,dashed,shape=SEbox,inner sep=2pt,minimum height=6mm,fill=white]
\tikzstyle{ddmaptrans}=[draw,doubled,dashed,shape=SWbox,inner sep=2pt,minimum height=6mm,fill=white]
\tikzstyle{ddmapconj}=[draw,doubled,dashed,shape=NWbox,inner sep=2pt,minimum height=6mm,fill=white]

\boxshape{sNEbox}{0pt}{3pt}
\boxshape{sSEbox}{0pt}{-3pt}
\boxshape{sNWbox}{3pt}{0pt}
\boxshape{sSWbox}{-3pt}{0pt}
\tikzstyle{smap}=[draw,shape=sNEbox,fill=white]
\tikzstyle{smapdag}=[draw,shape=sSEbox,fill=white]
\tikzstyle{smapadj}=[draw,shape=sSEbox,fill=white]
\tikzstyle{smaptrans}=[draw,shape=sSWbox,fill=white]
\tikzstyle{smapconj}=[draw,shape=sNWbox,fill=white]

\tikzstyle{dsmap}=[draw,dashed,shape=sNEbox,fill=white]
\tikzstyle{dsmapdag}=[draw,dashed,shape=sSEbox,fill=white]
\tikzstyle{dsmaptrans}=[draw,dashed,shape=sSWbox,fill=white]
\tikzstyle{dsmapconj}=[draw,dashed,shape=sNWbox,fill=white]

\boxshape{mNEbox}{0pt}{10pt}
\boxshape{mSEbox}{0pt}{-10pt}
\boxshape{mNWbox}{10pt}{0pt}
\boxshape{mSWbox}{-10pt}{0pt}
\tikzstyle{mmap}=[draw,shape=mNEbox]
\tikzstyle{mmapdag}=[draw,shape=mSEbox]
\tikzstyle{mmaptrans}=[draw,shape=mSWbox]
\tikzstyle{mmapconj}=[draw,shape=mNWbox]

\tikzstyle{mmapgray}=[draw,fill=gray!40!white,shape=mNEbox]
\tikzstyle{smapgray}=[draw,fill=gray!40!white,shape=sNEbox]

\makeatletter
\pgfdeclareshape{cornerpoint}{
\inheritsavedanchors[from=rectangle] 
\inheritanchorborder[from=rectangle]
\inheritanchor[from=rectangle]{center}
\inheritanchor[from=rectangle]{north}
\inheritanchor[from=rectangle]{south}
\inheritanchor[from=rectangle]{west}
\inheritanchor[from=rectangle]{east}
\backgroundpath{
\southwest \pgf@xa=\pgf@x \pgf@ya=\pgf@y
\northeast \pgf@xb=\pgf@x \pgf@yb=\pgf@y

\pgfmathsetmacro{\pgf@shorten@left}{\pgfkeysvalueof{/tikz/shorten left}}
\pgfmathsetmacro{\pgf@shorten@right}{\pgfkeysvalueof{/tikz/shorten right}}

\pgfpathmoveto{\pgfpoint{0.5 * (\pgf@xa + \pgf@xb)}{\pgf@ya - 5pt}}
\pgfpathlineto{\pgfpoint{\pgf@xa - 8pt + \pgf@shorten@left}{\pgf@yb - 1.5 * \pgf@shorten@left}}
\pgfpathlineto{\pgfpoint{\pgf@xa - 8pt + \pgf@shorten@left}{\pgf@yb}}
\pgfpathlineto{\pgfpoint{\pgf@xb + 8pt - \pgf@shorten@right}{\pgf@yb}}
\pgfpathlineto{\pgfpoint{\pgf@xb + 8pt - \pgf@shorten@right}{\pgf@yb - 1.5 * \pgf@shorten@right}}
\pgfpathclose
}
}

\pgfdeclareshape{cornercopoint}{
\inheritsavedanchors[from=rectangle] 
\inheritanchorborder[from=rectangle]
\inheritanchor[from=rectangle]{center}
\inheritanchor[from=rectangle]{north}
\inheritanchor[from=rectangle]{south}
\inheritanchor[from=rectangle]{west}
\inheritanchor[from=rectangle]{east}
\backgroundpath{
\southwest \pgf@xa=\pgf@x \pgf@ya=\pgf@y
\northeast \pgf@xb=\pgf@x \pgf@yb=\pgf@y

\pgfmathsetmacro{\pgf@shorten@left}{\pgfkeysvalueof{/tikz/shorten left}}
\pgfmathsetmacro{\pgf@shorten@right}{\pgfkeysvalueof{/tikz/shorten right}}

\pgfpathmoveto{\pgfpoint{0.5 * (\pgf@xa + \pgf@xb)}{\pgf@yb + 5pt}}
\pgfpathlineto{\pgfpoint{\pgf@xa - 8pt + \pgf@shorten@left}{\pgf@ya + 1.5 * \pgf@shorten@left}}
\pgfpathlineto{\pgfpoint{\pgf@xa - 8pt + \pgf@shorten@left}{\pgf@ya}}
\pgfpathlineto{\pgfpoint{\pgf@xb + 8pt - \pgf@shorten@right}{\pgf@ya}}
\pgfpathlineto{\pgfpoint{\pgf@xb + 8pt - \pgf@shorten@right}{\pgf@ya + 1.5 * \pgf@shorten@right}}
\pgfpathclose
}
}

\pgfdeclareshape{langpoint}{
\inheritsavedanchors[from=rectangle] 
\inheritanchorborder[from=rectangle]
\inheritanchor[from=rectangle]{center}
\inheritanchor[from=rectangle]{north}
\inheritanchor[from=rectangle]{south}
\inheritanchor[from=rectangle]{west}
\inheritanchor[from=rectangle]{east}
\backgroundpath{
\southwest \pgf@xa=\pgf@x \pgf@ya=\pgf@y
\northeast \pgf@xb=\pgf@x \pgf@yb=\pgf@y

\pgfmathsetmacro{\pgf@shorten@left}{\pgfkeysvalueof{/tikz/shorten left}}
\pgfmathsetmacro{\pgf@shorten@right}{\pgfkeysvalueof{/tikz/shorten right}}

\pgfpathmoveto{\pgfpoint{0.5 * (\pgf@xa + \pgf@xb)}{\pgf@ya - 2pt}}
\pgfpathlineto{\pgfpoint{\pgf@xa - 8pt}{\pgf@yb - 3 * \pgf@shorten@left + 5pt}} 
\pgfpathlineto{\pgfpoint{\pgf@xa - 8pt}{\pgf@yb -1pt}}
\pgfpathlineto{\pgfpoint{\pgf@xb + 8pt}{\pgf@yb -1pt}}
\pgfpathlineto{\pgfpoint{\pgf@xb + 8pt}{\pgf@yb - 3 * \pgf@shorten@left + 5pt}}
\pgfpathclose
}
}

\pgfdeclareshape{langcopoint}{
\inheritsavedanchors[from=rectangle] 
\inheritanchorborder[from=rectangle]
\inheritanchor[from=rectangle]{center}
\inheritanchor[from=rectangle]{north}
\inheritanchor[from=rectangle]{south}
\inheritanchor[from=rectangle]{west}
\inheritanchor[from=rectangle]{east}
\backgroundpath{
\southwest \pgf@xa=\pgf@x \pgf@ya=\pgf@y
\northeast \pgf@xb=\pgf@x \pgf@yb=\pgf@y

\pgfmathsetmacro{\pgf@shorten@left}{\pgfkeysvalueof{/tikz/shorten left}}
\pgfmathsetmacro{\pgf@shorten@right}{\pgfkeysvalueof{/tikz/shorten right}}

\pgfpathmoveto{\pgfpoint{0.5 * (\pgf@xa + \pgf@xb)}{\pgf@yb +0pt}}
\pgfpathlineto{\pgfpoint{\pgf@xa - 8pt}{\pgf@ya + 3 * \pgf@shorten@left - 5pt}} 
\pgfpathlineto{\pgfpoint{\pgf@xa - 8pt}{\pgf@ya + 1pt}}
\pgfpathlineto{\pgfpoint{\pgf@xb + 8pt}{\pgf@ya + 1pt}}
\pgfpathlineto{\pgfpoint{\pgf@xb + 8pt}{\pgf@ya + 3 * \pgf@shorten@left - 5pt}}
\pgfpathclose
}
}

\pgfdeclareshape{langrect}{
\inheritsavedanchors[from=rectangle] 
\inheritanchorborder[from=rectangle]
\inheritanchor[from=rectangle]{center}
\inheritanchor[from=rectangle]{north}
\inheritanchor[from=rectangle]{south}
\inheritanchor[from=rectangle]{west}
\inheritanchor[from=rectangle]{east}
\backgroundpath{
\southwest \pgf@xa=\pgf@x \pgf@ya=\pgf@y
\northeast \pgf@xb=\pgf@x \pgf@yb=\pgf@y

\pgfmathsetmacro{\pgf@shorten@left}{\pgfkeysvalueof{/tikz/shorten left}}
\pgfmathsetmacro{\pgf@shorten@right}{\pgfkeysvalueof{/tikz/shorten right}}

\pgfpathmoveto{\pgfpoint{\pgf@xa - 8pt}{\pgf@ya + 3 * \pgf@shorten@left - 5pt}} 
\pgfpathlineto{\pgfpoint{\pgf@xa - 8pt}{\pgf@ya + 1pt}}
\pgfpathlineto{\pgfpoint{\pgf@xb + 8pt}{\pgf@ya + 1pt}}
\pgfpathlineto{\pgfpoint{\pgf@xb + 8pt}{\pgf@ya + 3 * \pgf@shorten@left - 5pt}}
\pgfpathclose
}
}

\makeatother

\pgfkeyssetvalue{/tikz/shorten left}{0pt}
\pgfkeyssetvalue{/tikz/shorten right}{0pt}

\tikzstyle{kpoint common}=[draw,fill=white,inner sep=1pt,minimum height=4mm]

\tikzstyle{langstate}=[shape=langcopoint,shorten left=5pt,kpoint common,font=\footnotesize]
\tikzstyle{langeffect}=[shape=langpoint,shorten left=5pt,kpoint common,font=\footnotesize]
\tikzstyle{langstatedash}=[shape=langcopoint,dashed, shorten left=5pt,kpoint common,font=\footnotesize]
\tikzstyle{langeffectdash}=[shape=langpoint,dashed, shorten left=5pt,kpoint common,font=\footnotesize]
\tikzstyle{langbox}=[shape=langrect,shorten left=5pt,kpoint common,font=\footnotesize] 

\tikzstyle{kpoint}=[shape=cornerpoint,shorten left=5pt,kpoint common]
\tikzstyle{kpoint adjoint}=[shape=cornercopoint,shorten left=5pt,kpoint common]

\tikzstyle{kpoint conjugate}=[shape=cornerpoint,shorten right=5pt,kpoint common]
\tikzstyle{kpoint transpose}=[shape=cornercopoint,shorten right=5pt,kpoint common]
\tikzstyle{kpoint symm}=[shape=cornerpoint,shorten left=5pt,shorten right=5pt,kpoint common]

\tikzstyle{black kpoint}=[shape=cornerpoint,shorten left=5pt,kpoint common,fill=black,font=\color{white}]
\tikzstyle{black kpoint adjoint}=[shape=cornercopoint,shorten left=5pt,kpoint common,fill=black,font=\color{white}]
\tikzstyle{black kpointadj}=[shape=cornercopoint,shorten left=5pt,kpoint common,fill=black,font=\color{white}]

\tikzstyle{black dkpoint}=[shape=cornerpoint,shorten left=5pt,kpoint common,fill=black, doubled,font=\color{white}]
\tikzstyle{black dkpoint adjoint}=[shape=cornercopoint,shorten left=5pt,kpoint common,fill=black, doubled,font=\color{white}]
\tikzstyle{black dkpointadj}=[shape=cornercopoint,shorten left=5pt,kpoint common,fill=black, doubled,font=\color{white}]

\tikzstyle{kpointdag}=[kpoint adjoint]
\tikzstyle{kpointadj}=[kpoint adjoint]
\tikzstyle{kpointconj}=[kpoint conjugate]
\tikzstyle{kpointtrans}=[kpoint transpose]

\tikzstyle{big kpoint}=[kpoint, minimum width=1.2 cm, minimum height=8mm, inner sep=4pt, text depth=3mm]

\tikzstyle{wide kpoint}=[kpoint, minimum width=1 cm, inner sep=2pt]
\tikzstyle{wide kpointdag}=[kpointdag, minimum width=1 cm, inner sep=2pt]
\tikzstyle{wide kpointconj}=[kpointconj, minimum width=1 cm, inner sep=2pt]
\tikzstyle{wide kpointtrans}=[kpointtrans, minimum width=1 cm, inner sep=2pt]

\tikzstyle{gray kpoint}=[kpoint,fill=gray!50!white]
\tikzstyle{gray kpointdag}=[kpointdag,fill=gray!50!white]
\tikzstyle{gray kpointadj}=[kpointadj,fill=gray!50!white]
\tikzstyle{gray kpointconj}=[kpointconj,fill=gray!50!white]
\tikzstyle{gray kpointtrans}=[kpointtrans,fill=gray!50!white]

\tikzstyle{gray dkpoint}=[kpoint,fill=gray!50!white,doubled]
\tikzstyle{gray dkpointdag}=[kpointdag,fill=gray!50!white,doubled]
\tikzstyle{gray dkpointadj}=[kpointadj,fill=gray!50!white,doubled]
\tikzstyle{gray dkpointconj}=[kpointconj,fill=gray!50!white,doubled]
\tikzstyle{gray dkpointtrans}=[kpointtrans,fill=gray!50!white,doubled]

\tikzstyle{white label}=[draw,fill=white,rectangle,inner sep=0.7 mm]
\tikzstyle{gray label}=[draw,fill=gray!50!white,rectangle,inner sep=0.7 mm]
\tikzstyle{black label}=[draw,fill=black,rectangle,inner sep=0.7 mm]

\tikzstyle{dkpoint}=[kpoint,doubled]
\tikzstyle{wide dkpoint}=[wide kpoint,doubled]
\tikzstyle{dkpointdag}=[kpoint adjoint,doubled]
\tikzstyle{wide dkpointdag}=[wide kpointdag,doubled]
\tikzstyle{dkcopoint}=[kpoint adjoint,doubled]
\tikzstyle{dkpointadj}=[kpoint adjoint,doubled]
\tikzstyle{dkpointconj}=[kpoint conjugate,doubled]
\tikzstyle{dkpointtrans}=[kpoint transpose,doubled]

\tikzstyle{kscalar}=[kpoint common, shape=EBox, inner xsep=-1pt, inner ysep=3pt,font=\small]
\tikzstyle{kscalarconj}=[kpoint common, shape=WBox, inner xsep=-1pt, inner ysep=3pt,font=\small]

 \tikzstyle{upground}=[circuit ee IEC,ground,rotate=90,scale=2.5]
 \tikzstyle{downground}=[circuit ee IEC,ground,rotate=-90,scale=2.5]
 \tikzstyle{bigground}=[regular polygon,regular polygon sides=3,draw=gray,scale=0.50,inner sep=-0.5pt,minimum width=10mm,fill=gray]

\tikzstyle{arrs}=[-latex,font=\small,auto]
\tikzstyle{arrow plain}=[arrs]
\tikzstyle{arrow dashed}=[dashed,arrs]
\tikzstyle{arrow bold}=[very thick,arrs]
\tikzstyle{arrow hide}=[draw=white!0,-]
\tikzstyle{arrow reverse}=[latex-]
\tikzstyle{cdnode}=[]

\tikzstyle{H}=[-, style=dashed]
\tikzstyle{K}=[-, line width=1pt]
\tikzstyle{Kv}=[-, line width=1pt, ->]
\tikzstyle{Kv<>}=[-,line width=1pt,{<->}]
\tikzstyle{gF}=[-, draw=none, fill={rgb,255: red,191; green,191; blue,191}]
\tikzstyle{KB}=[-, draw=blue, line width=1pt]
\tikzstyle{KO}=[-, draw={rgb,255: red,255; green,128; blue,0}, line width=1pt]
\tikzstyle{KL}=[-, draw={rgb,255: red,191; green,255; blue,0}, line width=1pt]
\tikzstyle{KHO}=[-, draw={rgb,255: red,255; green,128; blue,0}, style=dashed, line width=1pt]
\tikzstyle{KTG}=[-, draw={rgb,255: red,128; green,128; blue,128}, style=dotted, line width=1pt]
\tikzstyle{KTlG}=[-, draw={rgb,255: red,191; green,191; blue,191}, style=dotted, line width=1pt]
\tikzstyle{KBv}=[-, draw=blue, ->]
\tikzstyle{KOv}=[-, draw={rgb,255: red,255; green,128; blue,0}, line width=1pt, ->]
\tikzstyle{KLv}=[-, draw={rgb,255: red,191; green,255; blue,0}, ->]
\tikzstyle{T}=[-, style=dotted]
\tikzstyle{wF}=[-, fill=white, draw=none]
\tikzstyle{KH}=[-, style=dashed, line width=1pt]
\tikzstyle{Hv}=[->, style=dashed]
\tikzstyle{cv}=[-,right hook->]
\tikzstyle{vv}=[-,->>]
\tikzstyle{v}=[-,->]
\tikzstyle{<>}=[-,<->]
\tikzstyle{Hvv}=[-,->>,style=dashed]
\tikzstyle{Kvp}=[->]
\tikzstyle{KTB}=[-, draw=blue, style=dotted, line width=1pt]
\tikzstyle{KBgF}=[-, fill={rgb,255: red,191; green,191; blue,191}, draw=blue, line width=1 pt]
\tikzstyle{KBggF}=[-, fill={rgb,255: red,128; green,128; blue,128}, draw=blue, line width=1 pt]
\tikzstyle{b-wf}=[-, fill=white]
\tikzstyle{b-gf}=[-, fill={rgb,255: red,191; green,191; blue,191}, draw=black]
\tikzstyle{KHB}=[-, draw=blue, style=dashed, line width=1pt]

\tikzstyle{bigunit}=[dot,fill=white,text depth=-0.2mm]
\tikzstyle{smallblackdot}=[fill=black, inner sep=0mm,minimum width=1mm,minimum height=1mm,draw,shape=circle]
\tikzstyle{smallorangedot}=[fill={rgb,255: red,255; green,128; blue,0}, inner sep=0mm,minimum width=1mm,minimum height=1mm,draw,shape=circle]
\tikzstyle{smallbluedot}=[fill=blue, inner sep=0mm,minimum width=1mm,minimum height=1mm,draw,shape=circle]

\definecolor{hexcolor0xa9a9a9}{rgb}{0.663,0.663,0.663} 
\tikzstyle{GrayLine}=[dashed,draw=hexcolor0xa9a9a9] 
\tikzstyle{gray}=[dashed,draw=hexcolor0xa9a9a9]

\theoremstyle{definition}
\newtheorem{theorem}{Theorem}[section]
\newtheorem*{theorem*}{Theorem}
\newtheorem{corollary}[theorem]{Corollary}
\newtheorem{lemma}[theorem]{Lemma} 
\newtheorem{proposition}[theorem]{Proposition}

\newtheorem{example}[theorem]{Example} 

\newtheorem{example*}[theorem]{Example*}
\newtheorem{examples*}[theorem]{Examples*}

\newtheorem{remark*}[theorem]{Remark*}

\newtheorem{convention}[theorem]{Convention}

\newtheorem{disclaimer}[theorem]{Disclaimer}
\newtheorem{refinement}[theorem]{Refinement}
\newtheorem{thesis}[theorem]{Thesis}

\def\bR{\begin{color}{red}}  
\def\bB{\begin{color}{blue}}
\def\bM{\begin{color}{magenta}}  
\def\bC{\begin{color}{cyan}}
\def\bW{\begin{color}{white}}
\def\bBl{\begin{color}{black}}
\def\bG{\begin{color}{green}}
\def\bY{\begin{color}{yellow}}
\def\e{\end{color}\xspace}
\newcommand{\bit}{\begin{itemize}}
\newcommand{\eit}{\end{itemize}\par\noindent}
\newcommand{\ben}{\begin{enumerate}}
\newcommand{\een}{\end{enumerate}\par\noindent}
\newcommand{\beq}{\begin{equation}}
\newcommand{\eeq}{\end{equation}\par\noindent}
\newcommand{\beqa}{\begin{eqnarray*}}
\newcommand{\eeqa}{\end{eqnarray*}\par\noindent}
\newcommand{\beqn}{\begin{eqnarray}}
\newcommand{\eeqn}{\end{eqnarray}\par\noindent}

\title{Distilling Text into Circuits}  

\author{Vincent Wang-Ma\'{s}cianica${}^{\dagger\ddagger}$, Jonathon Liu${}^{\dagger}$ and Bob Coecke${}^{\dagger}$\\ \ \\ 
${}^{\dagger}$Quantinuum, Quantum Compositional Intelligence, Oxford\\
${}^{\ddagger}$Oxford University, Department of Computer Science\\ 
}     
  
\begin{document}            
\maketitle  
\begin{abstract}    
This paper concerns the structure of meanings within natural language.  In \cite{CoeckeText} a framework for natural language named DisCoCirc was sketched  that has the following features:
\ben
\item[(1)] it is both compositional and distributional (a.k.a.~vectorial);  
\item[(2)] it applies to general text, not just sentences; 
\item[(3)] it captures linguistic `connections' between meanings (cf.~grammar);   
\item[(4)] word meanings get updated as text progresses; 
\item[(5)] sentence types reflect the words that have their meaning updated;  
\item[(6)] language ambiguity is naturally accommodated. 
\een
In this paper, we realise DisCoCirc for a substantial fragment of English. 

We moreover show that when passing to DisCoCirc's \underline{text circuits}, some `grammatical bureaucracy' is eliminated.
That is, DisCoCirc  displays a significant degree of:
\ben
\item[(7)] inter- and intra-language independence.  
\een
By inter-language independence we mean for example independence from word-order conventions that differ across languages, and by intra-language independence we mean for example independence from choices like using many short sentences versus few long sentences. 
This inter-language independence means our text circuits should carry over to other languages, unlike the language-specific typings of categorial grammars.
Hence, text circuits are a lean structure for the `actual substance of text', that is, the inner-workings of meanings within text across several layers of expressiveness (cf.~words, sentences, text), and  may capture that what is truly universal beneath grammar. 

The elimination of grammatical bureaucracy also explains why DisCoCirc:
\bit  
\item[(8)]  applies beyond language, e.g.~to spatial \cite{TalkSpace}, visual \cite[\S 4.7]{CoeckeText} and other cognitive modes.  
\eit
Indeed, while humans could not verbally communicate in terms of text circuits, machines can, and abstract text circuits can be conceived as linguistic structure tailored for machines.

Concretely, we begin by defining a `hybrid grammar' for a fragment of English - this is a purpose-built, minimal grammatical formalism needed to obtain text circuits. We then detail a translation process such that all text generated by this grammar yields a text circuit:
\[
\scalebox{0.60}{$\input{bigex-0.tikz} \qquad\ \ \ \longrightarrow\qquad \ \ \ \ \input{bigex-circuit.tikz}$}
\]
Conversely, for any text circuit obtained by freely composing the generators, there exists a text (with hybrid grammar) that gives rise to it. Hence:
\ben
\item[(9)] text circuits are generative for text.  
\een
%
\end{abstract}   
 
\tableofcontents

\newpage
\section{Introduction}\label{sec:intoduction}       
  
This paper concerns a compositional framework in which distributional (a.k.a.~vectorial) meanings and linguistic structure  meaningfully  as well as transparently  interact. It is moreover flexible enough to accommodate compositionality beyond Frege's bottom-up notion \cite{compositionality}. For example, in natural language, local meanings of words are often informed by the global textual context. Also, modern machine learning methods involve such top-down meaning flows, in the sense that global training enforces the states of local neurons -- although their meanings are typically less clear.
  
A first model combining distributional meanings with grammar -- in this particular case, categorial grammar \cite{Ajdukiewicz, Lambek0, steedman1987combinatory, Lambek1} -- was DisCoCat, proposed in 2008  \cite{CSC}. DisCoCat enjoyed empirical support \cite{baroni2010nouns, GrefSadr, KartSadr, wijnholds-sadrzadeh-2019-evaluating, MarthaNeural, wijnholds2020representation}, and was an attempt to unite compositional and data-driven approaches in natural language processing, after they had gone their separate ways \cite{Gazdar, ClarkPulman, pereira_formal_2000}. In particular, accounting for grammar makes many `linguistic connections' between words and phrases explicit, so they no longer need to be learned when using machine learning methods.\footnote{Note here  in particular that DisCoCat and DisCoCirc naturally handle ambiguity concerning meaning, grammar or both, using density matrices and corresponding superoperators \cite{calco2015,  kartsaklis2015compositional, RobinMSc, MarthaNeural}.}
  
More recently, DisCoCirc \cite{CoeckeText} was proposed as an improvement of DisCoCat, addressing three major shortcomings. Firstly, DisCoCat  produced meanings of phrases or sentences given word meanings, but  not text meanings from sentence meanings.  Secondly, in DisCoCat all meanings were static, while meanings of words may change as text progresses. In fact, this shortcoming also applies to most current large language models. Thirdly, in DisCoCat the space of sentence meanings was unspecified, hence non-canonical. In DisCoCirc it is made up from the relevant nouns.

The manner in which DisCoCirc realises these improvements, is by letting text take the form of \underline{text circuits}, the two-dimensionality of which is critical to realising each of the above stated improvements.  We discuss this dimensionality issue separately below. 

The `baby version' of DisCoCirc in \cite{CoeckeText} has been applied to a number of problems including spatio-temporal models of language meaning \cite{TalkSpace}, logical and conversational negation in natural language \cite{MarthaNeg, rodatz2021conversational, shaikh2021composing}, solving logical puzzles \cite{semspace_tiffany_2020, duneau2021parsing}, and in particular, quantum natural language processing (QNLP) where text circuits match the machine-level code of quantum hardware \cite{QNLP-foundations}.\footnote{The software tool kit {\tt lambeq} for doing QNLP implementations  \cite{Nature, QNLPPlus100} has recently been made available \cite{kartsaklis2021lambeq}. However, in its present form DisCoCirc is not part of it yet, but this will change in the near future.} There also have been some further theoretical elaborations on DisCoCirc \cite{CoeckeMeich,   hefford2020categories, GemmaMartha, GramEqs}.    

However, the proposal in \cite{CoeckeText} for DisCoCirc was restricted to very basic SVO-sentences, possibly with a relative pronoun. In these cases, given the (pregroup) grammar of a sentence \cite{Lambek1, LambekBook}, a corresponding circuit was produced that could then be further composed into larger text. But there was no general recipe to produce text circuits from general text, and consequently, no characterisation of what a `valid' text circuit is.  We do all of that in this paper, and we also establish the precise relationship between text and text circuits. Some immediate hurdles that one encounters are the following:
\bit 
\item Firstly, there is no general agreement of what the theory of grammar actually is, as witnessed by the fact that there are several entirely different theories available. To provide just a sample, there are: typelogical grammars such as combinatory categorial grammar \cite{steedman1987combinatory}, Lambek grammar \cite{Lambek0} and pregroup grammar \cite{LambekBook}; labelled-graph approaches such as dependency \cite{tesniere_elements_2015} and link \cite{sleator_parsing_1995} grammar; box-nesting approaches such as discourse representation theory \cite{kamp_discourse_2010} and \cite{bresnan_bresnan_2000}; and generative approaches of various shapes, such as transformational grammar \cite{radford_transformational_1988}, tree-adjoining grammar \cite{zwicky_tree_1985} etc.  
\item Secondly,  the nature of natural language makes concrete and complete mathematical descriptions and algorithms difficult, so in practice one relies on statistical parsers. For example, to concretely specify a parsing algorithm for dependency grammar prohibitively requires a complete catalogue of every kind of grammatical relationship possible in a language, so the next best thing is a statistical parser such as SpaCy \cite{spacy2}.
\item Thirdly, in order to produce text circuits, pronouns need to be resolved, and scope needs to be specified, as these are also crucial to identifying basic meaning connectedness.  
\item Finally, grammar typically only deals with sentences, not text.   
\eit

To address these issues, in Section \ref{sec:grammar}  we outline a purpose-built `hybrid grammar' for text.
Our goal here is not to establish a full-blown new grammar theory, but rather build just enough grammatical structure so that we can establish a bridge between text equipped with grammar and text circuits, for a reasonably large fragment of language.
To build this minimal grammar, we use a phrase structure grammar augmented with a number of extra ingredients: `pronominal links' that identify reoccurring referents, `phrase regions' that identify the scope of words that take a sentential complement, and `transformational grammar rules' that let us form more complex sentences and introduce relative pronouns.

In Section \ref{sec:congraphas}, we translate text with hybrid grammar into string diagrams, so obtaining the intermediate structure of \underline{text diagrams}, which incorporate the grammatical phenomena of our hybrid grammar -- phrase structure, pronominal links, phrase scope, and transformational grammar rules  -- into a single mathematical structure. Artefacts required to handle transformational grammar rules in our grammar vanish in the setting of text diagrams.
In Section \ref{sec:circs}, we introduce text circuits, which are generative for text, and show how to go from text diagrams to text circuits.
In Section \ref{sec:thm} we prove our main result that characterises text circuits as the `essential meaning connectedness' of text equipped with hybrid grammar, and we further outline how the theory of text circuits may be conservatively extended to accommodate more features of language. We show that every text of hybrid grammar yields a text circuit, and we also show that every text circuit arises from some text. 

Interestingly, the map from text equipped with grammar to text circuits is a proper surjection, that is, different texts may yield the same circuit. This reflects the fact that DisCoCirc eliminates what we refer to as `grammatical bureaucracy', abstracting away differences between texts that carry the same meanings. We discuss this feature separately below. Finally, in Section \ref{sec:discussion}, we elaborate on practical considerations such as parsing, expanding the fragment of language we model, as well as the relationships between text circuits and other grammatical and semantic formalisms. There is an important remark that follows from this comparison: although in this paper we have chosen to obtain text circuits from our hybrid grammar, it is clear that there are several other well-established formalisms that could be used as an alternative starting point for obtaining text circuits.

Altogether, we can represent the \underline{distillation} of text circuits from text as follows:
\[
\input{Distil1.tikz}  
\]
We explicitly construct this process and prove its essential properties, such as (proper) surjectivity.

However, this process assumes that we can unambiguously assign grammatical structure to the text, which is usually possible for short sentences, but not always for longer ones. We also want machines to be able to distil  circuits, and therefore in general distillation would rather look as follows:
\[
\input{Distil2.tikz}
\]
where \underline{parsing} is the process of automatically assigning grammatical and other structure to sentences and text. We won't explicitly develop the parsing step in this paper, as there are already many tools available for that purpose, for each of the ingredients our  grammar is made up from -- see Section \ref{sec:parsing}.
    
Our hybrid grammar for text is generative:
\[
\input{Generate1.tikz}
\]
where $*$ indicates the data that goes into the generation process.  However, following our results, we can now also generate text simply by composing circuits, and using the mapping that assigns each circuit to a text that generates it, here denoted by \underline{textualise}:
\[
\input{Generate2.tikz} 
\]
The fact that many texts yield the same text circuit, means that the resulting text comes in a normal form.  When working with text as circuits, as machines would do, then of course we can directly generate these without any further ado:
\[
\input{Generate3.tikz}
\]

We haven't mentioned yet how we can actually model meanings in our circuits. For DisCoCat, the analogue of the vector space embeddings of machine learning is the category of vector spaces, linear maps and tensor product, which is described in great detail in \cite{CatsII}.  Moreover, the flexibility of the framework allows one to move to density matrices and superoperators, which naturally accommodates lexical ambiguity  \cite{calco2015, kartsaklis2015compositional, RobinMSc, MarthaNeural}, and lexical entailment \cite{bankova2019graded, MarthaDot}.  The same model can also be used for DisCoCirc \cite[{\S} 6.1--6.3]{CoeckeText, CoeckeMeich, GemmaMartha}, although, following the results of this paper, with addition of higher-order operators.  We leave a detailed discussion of models to future writings.

\paragraph{Grammatical bureaucracy.}  What we mean in this paper by linguistic structure is not full-blown grammar, but rather just the minimal structure capturing `connectedness' of meanings. For example, in a sentence with a transitive verb, subject and object, both subject and object are connected to verb. In standard grammar, this structure is fixed by their relative ordering, but, every possible ordering of these three grammatical roles appears in some language in the world \cite{tomlin2014basic}. The conventions made within some language regarding word ordering is part of what we call \underline{inter-language grammatical bureaucracy}. Within a given language, there are other choices to be made when expressing meaning, like using a single long sentence versus many short sentences, the corresponding use of (relative) pronouns, and other stylistic aspects of texts, resulting in the fact that different texts may carry the same meanings. We call this  \underline{intra-language grammatical bureaucracy}. That different texts can share the same meaning induces equations between texts -- which elsewhere we called `grammar equations' \cite{GramEqs}. When passing to language circuits this grammatical bureaucracy gets eliminated, i.e.~we pass to equivalence classes, only retaining what could be called the `actual substance' of the text. Here are examples of different sentences or parts of text  resulting in the same language circuit:
\begin{center}
\begin{minipage}[b]{1\linewidth}
\[
\ \ \ \scalebox{0.75}{$
\begin{cases}
\texttt{SOBER ALICE WHO}\vspace{1.4mm}\\
\texttt{SEES DRUNK BOB}\vspace{1.4mm}\\
\texttt{CLUMSILY DANCE}\vspace{1.4mm}\\
\texttt{LAUGHS AT HIM.}
\end{cases}
$}\quad
\scalebox{0.75}{$
\begin{array}{l}
\begin{cases}
\texttt{ALICE SEES BOB DANCE CLUMSILY.}
\end{cases}\\
\begin{cases}
\texttt{ALICE LAUGHS AT BOB.} 
\end{cases}\\
\begin{cases}
\texttt{BOB IS DRUNK.}
\end{cases}\\
\begin{cases}
\texttt{ALICE IS SOBER.}
\end{cases}
\end{array}$
}\quad\!\!
\scalebox{0.75}{$
\begin{array}{l}
\begin{cases} 
\texttt{ALICE VOIT QUE BOB DANSER MALADROITEMENT.}
\end{cases}\\
\begin{cases}
\texttt{ALICE SE MOQUE DE BOB.}
\end{cases}\\
\begin{cases}
\texttt{ALICE EST SOBRE.}
\end{cases}\\
\begin{cases} 
\texttt{BOB EST IVRE.}
\end{cases}
\end{array}$
}
\]
\beq\label{circuitex}
\ \vspace{-8mm}
\eeq
\[
\scalebox{0.75}{\input{babeldev17.tikz}}
\]
\end{minipage}
\end{center}
An outline of inter-language-independence for the case of English and Urdu can be found in \cite{Urdu}.

Hence our text circuits may help answer the following question as well: 
\begin{center}
\fbox{\em What is truly universal beneath grammar?\em}  
\end{center} 
It has been claimed that grammar is hardwired in our human brains \cite{pinker_language_1995}, and a stronger claim is that there exists a unique mechanism that explains our capacity for language: a `universal grammar' \cite{Chomsky65}. All grammatical calculi throughout history, from P\-{a}nini to Montague, are in a way attempts to expose this `universality'.
However, there is a tension and a distance between the bureaucracy of grammar as captured by grammatical calculi, and our effortless use of language powered by a supposed `universal grammar'. If we can understand this bureaucracy -- not for the sake of faithfully photographing a language, but to remove the bureaucracy altogether -- then we may move closer to universality.

On the other hand, it is not so hard to see that this costly grammatical bureaucracy is quite hard to fully mathematically characterise. Standard natural language processing brute-forces the problem of grammatical bureaucracy by means of data-driven, machine learning style training of very large models, which may indeed be the most practical way, for now,  to handle these grammatical bureaucracies. Thus, in our view, generating general language, which includes the choice of language as well as the style, breaks down in two parts, namely,  building text circuits -- the bureaucracy-free actual substance of language -- and then turning text circuits into the actual text by reintroducing bureaucracy:      
\[
\input{generate.tikz} 
\]
where flags like \raisebox{-0.7mm}{\epsfig{figure=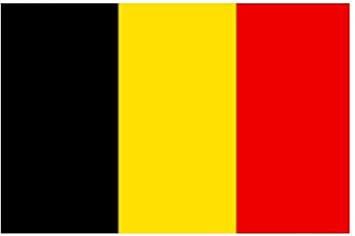,width=15pt}} refer to the specific language and the other icons like \raisebox{-1.3mm}{\epsfig{figure=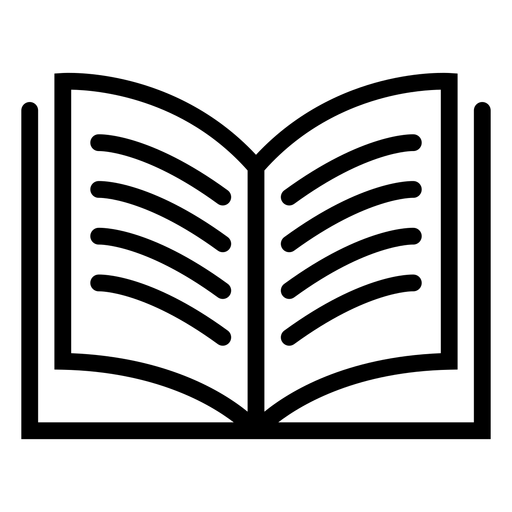,width=13pt}} to stylistic aspects.  

\paragraph{Two-dimensionality.} We believe that much of the grammatical bureaucracy is due to the fact that human language is a one-dimensional vehicle for higher-dimensional content, a view that mirrors Chomsky's minimalist program. So what do we mean by that? 
Sometimes things are forced to live in a world that is not their natural habitat. One such thing, in mathematics, are monoidal categories \cite{Benabou}, also known as tensor categories. This structure plays a central role in fields like quantum theory \cite{CKbook}, computer science \cite{AbrRetracing}, and many others, capturing the operations of parallel composition $\otimes$ (a.k.a.~`while')  and sequential composition $\circ$ (a.k.a.~`after').   Typically, in textbooks, this structure is represented by one-dimensional (1D) strings of algebraic symbols, and then, equations are needed to express their interaction, most notably,  so-called `bifunctoriality':   
\[
(g_1\otimes g_2)\circ(f_1\otimes f_2)=(g_1\circ f_1)\otimes(g_2\circ f_2) 
\]
On the other hand, representing it in terms of 2D diagrams, reading diagrams from top to bottom no equations are needed: 
\[
\left(\boxmap{g_1}\otimes \boxmap{g_2}\right)\circ\left(\boxmap{f_1}\otimes \boxmap{f_2}\right)
=
\left(\boxmap{g_1}\ \boxmap{g_2}\right)\circ\left(\boxmap{f_1}\ \boxmap{f_2}\right)
=\ 
\raisebox{0.6mm}{\input{twochain1.tikz}\  \input{twochain2.tikz}} 
\]
\[
\left(\boxmap{g_1}\circ \boxmap{f_1}\right)\otimes\left(\boxmap{g_2}\circ \boxmap{f_2}\right)
=
\left(\,\raisebox{0.6mm}{\input{twochain1.tikz}}\right)  \otimes \left(\,\raisebox{0.6mm}{\input{twochain2.tikz}}\right)
\ =\ 
\raisebox{0.6mm}{\input{twochain1.tikz}\  \input{twochain2.tikz}}   
\]
That is, by using the 2D format of diagrams, the defining symbolic equations are built-into the geometry of the plane.  The reason for the symbolic representation being more involved is that in order to force something 2D on a 1D line, artificial bureaucracy needs to be introduced (e.g.~bracketing) and this requires extra rules as well. A more detailed discussion of all of this can be found in \cite{CatsII, CKbook}.

Now, something similar is true for language, although in a somewhat more subtle manner, as indicated in diagram (\ref{circuitex}): placing words side-by-side creates different ways of saying the same thing, and this may also cause different conventions in different languages. Part of what this paper does is make the 1D vs.~2D argument formal in the case of language. 

A smaller example is the following, obtained via Google translate, where we now explicitly indicate the non-coinciding tree-grammar of the two sentences:
\[
\input{love.tikz}
\]
So different planar trees still give the same circuit. The reason for this is that while the trees look like a 2D mathematical entity, the words are still placed on a line, and the trees really behave as a bracketing structure for that line of words.

\paragraph{Compositional meanings.} The term compositionality has its origin within linguistics, and is usually associated with Frege's notion \cite{frege1914letter}, namely that the meaning of the whole is induced by (i) the meanings of the parts, and, (ii) how these parts fit together. Concretely: the meaning of a sentence arises from: (i) the meaning of its words, and, (ii) the grammatical structure of the sentence. As has been detailed in \cite{nefdt2020puzzle, compositionality} this conception of compositionality cannot be maintained, and a more general notion of compositionality has been laid out in \cite{compositionality}. Our text circuits are an example of that more general notion of compositionality, where also the meaning of the whole can inform the meanings of the parts.  In other words, compositionality constitutes a relationship across scales in which smaller and larger parts participate on equal footing. We refer the reader to \cite{qspeak} for a complementary discussion. 
The particular compositional form that language takes when moving to text circuits, moreover applies beyond language, e.g.~the spatial \cite{TalkSpace}, the visual \cite[\S 4.7]{CoeckeText} and the same is true for other cognitive modes \cite{ConcSpacI}.  In that sense, in the form of   text circuits, language appears to be some kind of `compositional logic of the world'.

\section{A hybrid grammar for text}\label{sec:grammar}

We now introduce our hybrid grammar, which is generative and aims to capture linguistic connectedness. 
We develop it in three steps. First, we start with a `context-sensitive'\footnote{Our grammar may be presented as a context-free one, but we choose a context-sensitive presentation for an easier mathematical connection to text diagrams and text circuits. There are only two rules in our hybrid grammar that are \emph{potentially} context-sensitive: the rule for introducing adpositions of transitive verbs, and the rules for managing phrase scope. Both may be recast in context-free form, which we sketch in footnotes \ref{footnote:cfarg} and \ref{footnote:cfarg2}. We refer to reader to \cite{hopcroft_introduction_1979} for definitions of context-free and context-sensitive grammars.} 
grammar for simple sentences that only contain one verb.
Second, we introduce the notion of pronominal links, which identify recurring nouns, and pronouns with their referents.
Further, we will introduce rules that allow us to fuse together simple sentences with recurring nouns via relative pronouns.
Thirdly, we introduce the notion of verbs that accept a sentential complement and phrase scope boundary. This expands our fragment to deal with compound sentences containing subphrases that are themselves sentences.

\begin{disclaimer}
As previously noted, our aim is to build a minimal grammar that allows us to generate text circuits from text and state crisp mathematical results. To this end, we use a Frankensteinesque hybrid of basic ideas from different formalisms: we use Chomsky's transformational phrase structure grammars, adjusted with features of Lambek's pregroups; pronominal links are inspired by discourse representation theory, and phrase boundaries are inspired by dependency grammars. Section \ref{sec:discussion} outlines these relationships in more detail.
\end{disclaimer}

\begin{disclaimer}
For the sake of clarity, we do not deal with some grammatical phenomena (like tense), omit certain grammatical patterns (we assume adverbs always come before the verb), and only deal with part of language (we do not consider determiners and quantifiers here, and we assume that ditransitive verbs have equivalent presentations as transitive verbs with adpositions.) This approach also has the usual shortcomings of formal grammars -- such as not taking into account adjective order for purpose, origin, etc. in languages like English -- that can be dealt with in the usual ways, requiring grammar to be mixed up with meanings. In fact, DisCoCat/DisCoCirc are all about combining grammar and meaning, and we (as many others) believe that ultimately they shouldn't exist independently, but should mutually inform each other. Practical NLP has empirically shown that this is essential for producing efficient tools.
\end{disclaimer}

\begin{disclaimer}	
Pronominal link and phrase boundary data are not uniquely determined by text when given as just a string of words, without any further context -- but then again, neither are grammatical types in many cases. The reason for including them is their necessity for obtaining \emph{disambiguated} text structure which we can reason about mathematically. 
\end{disclaimer}

\subsection{Preliminaries}

We model \emph{phrase structure} -- the structure that constrains the order of words in a sentence -- by providing a 
\emph{string rewrite system} that 
consists of (a usually finite collection of) \emph{production rules} \cite{hopcroft_introduction_1979}:
\[  
\alpha \mapsto \beta
\]
where $\alpha$ and $\beta$ are strings of symbols.  In our case, they may be phrase components, such as \texttt{NP}, or English words, such as \texttt{\underline{BOB}}. We underline the latter in order to indicate that such symbols are \emph{terminal} i.e.~not rewriteable further by production rules.

All such grammars specify \emph{languages} -- a collection of strings of symbols -- as follows. A special \emph{start symbol} $\texttt{S}$ is specified, and the language associated with the rewrite-system is the collection of all strings of terminal symbols that can be produced by (finite) applications of the production rules available, for example, given rules:
\beqn\label{rule1}
\texttt{S} &\mapsto& \texttt{NP} \cdot \texttt{IVP}\\ \label{rule2}  
\texttt{NP} &\mapsto& \texttt{\underline{BOB}}\\ \label{rule3}
\texttt{IVP} &\mapsto& \texttt{\underline{DRINKS}}
\eeqn
where $\cdot$ is notation for string concatenation, we can produce simple sentences such as:
\beqa
\texttt{S} & \stackrel{(\ref{rule1})}{\mapsto} & \texttt{NP} \cdot \texttt{IVP}\\
&\stackrel{(\ref{rule2})}{\mapsto} & \texttt{\underline{BOB}} \cdot \texttt{IVP}\\
&\stackrel{(\ref{rule3})}{\mapsto} & \texttt{\underline{BOB}} \cdot \texttt{\underline{DRINKS}}
\eeqa
We will depict derivations of strings as planar ``trees". The diagrams are read from top to bottom:
\[
\input{ex.tikz}     
\]
These ``trees" may have multiple edges from a parent node to a child node.
\[
\input{AgBtBcopy.tikz} 
\]
We drop symbolic labels for intermediate symbols, and replace them by coloured edges. For example, \texttt{NP} becomes a black edge and \texttt{IVP} becomes a green edge. As we introduce the rules, we will also keep to a coloring convention for typed wires, such that later on we may omit typings such as \texttt{IVP} from diagrams without confusion.

\subsection{Simple sentence structure}    

We now introduce a phrase structure grammar by giving the tree-fragments for the grammatical types, initially for what we call \emph{simple} sentences, which have a single verb that does not take a sentential complement.

\paragraph{Intransitive and Transitive Verbs.}
A simple sentence may contain a single intransitive or transitive verb. In the former case, the sentence consists of a noun-phrase followed by a intransitive-verb-phrase (e.g. \texttt{\underline{ALICE RUNS.}}). In the latter case, a sentence consists of a noun-phrase, transitive-verb-phrase, and another noun-phrase (e.g.~\texttt{\underline{ALICE LIKES BOB.}}). For these the tree-fragments are as follows:
\[
\begin{tabular}{|c|c|} 
\hline
\text{Rule} & \text{Planar Tree-Fragment}  \\ \hline
$\texttt{S} \mapsto \texttt{NP} \cdot \texttt{IVP}$ & \input{IVcsg.tikz}  \\ \hline
$\texttt{S} \mapsto \texttt{NP}_1 \cdot \texttt{TVP} \cdot \texttt{NP}_2$ & \input{TVcsgcopy.tikz} \\ \hline
\end{tabular}
\]
There also are the terminal rules for verbs, where the terminal symbols of the grammar are  verbs of the appropriate type e.g.~intransitive or transitive:
\[
\begin{tabular}{|c|c|}
\hline
\text{Rule} & \text{Planar Tree-Fragment}  \\ \hline
$\texttt{IVP} \mapsto \texttt{\underline{IV}}$ & \input{IVcsgL.tikz} \\ \hline
$\texttt{TVP} \mapsto \texttt{\underline{TV}}$ & \input{TVcsgL.tikz} \\ \hline
\end{tabular}
\]
Going forward, we omit the terminal rules in favour of giving examples of finished derivations, from which terminals can be inferred.    
  
\paragraph{Adjectives.}
Adjectives can appear before a noun-phrase (e.g.~\texttt{\underline{DRUNK HAPPY BOB.}}) Or, using the copular \texttt{\underline{IS}} considered as a verb, a single adjective can appear after a noun-phrase (e.g.~\texttt{\underline{BOB IS DRUNK.}}). Tree fragments are as follows:  

\[
\begin{tabular}{|c|c|}
\hline
\text{Rule} & \text{Planar Tree-Fragment}  \\ \hline
$\texttt{NP} \mapsto \texttt{ADJ} \cdot \texttt{NP}$ & \input{ADJcsg.tikz} \\ \hline
$\texttt{S} \mapsto \texttt{NP} \cdot \texttt{\underline{IS}} \cdot \texttt{ADJ}$ &  \input{ADJiscsg.tikz}\\ \hline
\end{tabular}
\]

\paragraph{Adverbs.}
Adverbs can appear before a verb (e.g. \texttt{\underline{ALICE QUICKLY HAPPILY RUNS.}}):

\[
\begin{tabular}{|c|c|}
\hline
\text{Rule} & \text{Planar Tree-Fragment}  \\ \hline
$\texttt{IVP} \mapsto \texttt{ADV} \cdot \texttt{IVP}$ & \input{ADVIVcsg.tikz}  \\ \hline
$\texttt{TVP} \mapsto \texttt{ADV} \cdot \texttt{TVP}$ & \input{ADVTVcsg.tikz} \\ \hline
\end{tabular}
\]

\paragraph{Adpositions.}    
An adposition can appear to the right of an intransitive-verb-phrase, followed by a noun-phrase (e.g.~from \texttt{\underline{ALICE RUNS.}} to \texttt{\underline{ALICE RUNS TOWARDS BOB.}}). In the presence of a transitive-verb-phrase flanked by a noun-phrase to the right, we may add to the right an adposition followed by a noun-phrase (e.g.~from \texttt{\underline{ALICE THROWS BEER.}} to \texttt{\underline{ALICE THROWS BEER TOWARDS BOB.}}):\footnote{\label{footnote:cfarg} 
	We can present this context-sensitive rule in an equivalent context-free manner.
	Upon inspection of the rules of our hybrid grammar 
	one can see that $\texttt{TVP}$ always appears as part of a string $\texttt{TVP}\cdot\texttt{NP}$.
	Suppose we replace this string with a single new symbol \texttt{TVP'}
	which `expands' by the following new rule	
	$\texttt{TVP'} \mapsto \texttt{TVP} \cdot \texttt{NP}$.
	Then we can replace the `context-sensitive' rule with the context-free rule
	$\texttt{TVP'} \mapsto \texttt{TVP'} \cdot \texttt{ADP} \cdot \texttt{NP}$. 
	Additionally, we may replace all other rules involving \texttt{TVP} with a version involving \texttt{TVP'}, resulting in an equivalent context-free presentation.}
\[
\begin{tabular}{|c|c|}   
\hline
\text{Rule} & \text{Planar Tree-Fragment}  \\ \hline
$\texttt{IVP} \mapsto \texttt{IVP} \cdot \texttt{ADP} \cdot \texttt{NP}$ & \input{ADPIVcsg.tikz} \\ \hline
$\texttt{TVP} \cdot \texttt{NP}_1 \mapsto \texttt{TVP} \cdot \texttt{NP}_1 \cdot \texttt{ADP} \cdot \texttt{NP}_2$ & \input{ADPTVcsgcopy.tikz} \\ \hline
\end{tabular}  
\]

\begin{example} First applying the \texttt{TVP}-production rule we obtain $\texttt{NP} \cdot \texttt{TVP} \cdot \texttt{NP}$, which corresponds to the grammatical structure of \texttt{\underline{ALICE GIVES BEER}}. Then applying the $\texttt{TVP} \cdot \texttt{NP}$-production rule, we transform \texttt{\underline{GIVES BEER}} (i.e.~both a green and a black wire) into \texttt{\underline{GIVES BEER TO BOB}}. Altogether, we obtain the following ``tree", where the dashed line indicates the two ``subtrees": 
\[
\input{AgBtB.tikz} 
\]
\end{example}

\subsection{Compound sentence structure I: pronominal links}\label{sec:pronom-links}

Now we go beyond individual sentences. We call a sentence with more than one verb \emph{compound}. One way we obtain such sentences is by fusing simple sentences together using relative pronouns. We model the data of a pronoun using a \emph{pronominal link}, which identifies nouns from possibly different sentences. We depict these as arrows under the trees, pointing at identified nouns. For example, in the text: \texttt{\underline{ALICE IS SOBER.} \underline{ALICE GIVES BEER TO BOB.}}, we can identify the two occurrences of \texttt{\underline{ALICE}} as below:
\[
\input{AiSAgBtB.tikz}
\]
In the presence of a pronominal link, later occurrences of a noun are interchangeable with a pronoun that refers to an earlier occurrence:
\[
\input{AiSAgBtB2.tikz} 
\]


\paragraph{Subject relative pronouns.} When pronominal links occur in certain configurations, we may fuse the sentences of two or more parse trees together using \emph{relative pronouns}.
Firstly, \emph{subject relative pronouns} replaces the subject noun of a parse tree $\texttt{S}_2$, and points to a noun in another, previous parse tree $\texttt{S}_1$, usually the object noun. Diagrammatically:
\beq\label{transgram}
\input{srprule.tikz}  
\eeq
where the big triangle on the right hand side is a visual convention to indicate that the two sentences are `fused' as a single sentence.  A rewrite rule like (\ref{transgram}) is called a \emph{transformational grammar rule}. Transformations modify phrase structure trees. Below we encounter more examples of transformational rules for relative pronouns, and in Section \ref{sec:congraphas} we formally define what they are.

\begin{example}
For the text \texttt{\underline{ALICE LIKES BOB.}} \texttt{\underline{BOB HATES CLAIRE.}}
we start with two trees.  We identify the occurrences of \texttt{\underline{BOB}} with a pronominal link:
\[
\input{srpex1.tikz}
\]
We can now fuse the trees by replacing the second occurrence of \texttt{\underline{BOB}} with the relative pronoun \texttt{\underline{WHO}}, yielding:
\[
\input{srpex2.tikz}  
\]
\end{example}

In English there is another special use of the subject relative pronoun to coordinate a single noun across two subsequent phrases. Given parse trees corresponding to sentences $\texttt{S}_1$ and $\texttt{S}_2$ in that order, this special case arises when the subject noun of $\texttt{S}_2$ points towards a subject noun in $\texttt{S}_1$. The result of the tree-transformation is that we have, in order: the noun-phrase (along with any adjectives) of $\texttt{S}_1$, followed by $\texttt{S}_1$ with the later noun replaced by the relative pronoun, followed by $\texttt{S}_2$ with its pointing noun removed.  Using a multiarrow to point out more than two pronominally identified nouns, we depict this as follows:
\[
\input{ssrprule.tikz}
\]

\begin{example}
We revisit a previous example:
\[
\input{AiSAgBtB.tikz}
\]
By applying the special rule for subject relative pronouns, we  obtain the following sentence:
\[
\input{ssrpex.tikz}
\]
We have not encountered an isolated noun -- here indicated by \texttt{!} -- or blank labels \textvisiblespace \ before; these artefacts disappear in the setting of text diagrams that we introduce in Section \ref{sec:congraphas}. 
\end{example}  

\paragraph{Object Relative Pronouns.}  In consecutive sentences $\texttt{S}_1$ and $\texttt{S}_2$, if the object noun of $\texttt{S}_2$ points to the object noun of $\texttt{S}_1$, the object relative pronoun comes after the first occurrence, and the second occurrence of the noun is replaced by a blank:
\[
\input{orprule.tikz}
\]

\begin{example}  We start with the text \texttt{\underline{ALICE LIKES BOB.}} \texttt{\underline{CLAIRE GIVES BEER TO BOB.}}
\[\input{orpex1.tikz}\]
Applying the rule for object relative pronouns, we obtain: 
\[\input{orpex2.tikz}\]
\end{example}

\paragraph{Reflexive Pronouns}
Sometimes two nouns in the same simple sentence are pronominally the same. In these cases, we may introduce a reflexive pronoun.
\begin{example}
Here we have \texttt{ALICE} liking \texttt{ALICE}:
\[
\input{rprule.tikz}
\]
\end{example}

\begin{example}
We observe that in dealing with pronominal links for relative and reflexive pronouns in hybrid grammar, we encounter various constraints, which we show here. These constraints are explained by the structure of text diagrams.

Say we have the following text with pronominal links:
\[
\input{rporderex1.tikz}
\]
We can first merge on \texttt{\underline{ALICE}} using the subject relative pronoun rule, and then replace the second occurrence of \texttt{\underline{BOB}} with the regular pronoun \texttt{\underline{HIM}}\footnote{We assume for simplicity the pronoun takes whatever grammatical form is `correct', agreeing with e.g.~gender and number. We care mainly about the referential structure. } obtaining the grammatical:
\[
\input{rporderex2a.tikz}
\]
Or, we can first merge on \texttt{\underline{BOB}} using the object relative pronoun rule, and then replace the second occurrence of \texttt{\underline{ALICE}} with the regular pronoun \texttt{\underline{SHE}}, obtaining the grammatical:
\[
\input{rporderex2b.tikz}
\]
We cannot apply both the subject- and object- relative pronoun rules at once unless the sentences were initially separate, as this would yield the \bR \textbf{ungrammatical sentence}\e:
\[
\bR \input{rporderex2c.tikz} \e
\]
A similar phenomenon occurs for reflexive and relative pronouns. If we start with the text:
\[
\input{reforderex1.tikz}   
\]
We can obtain a grammatical sentence by first introducing a reflexive pronoun, and then using the subject relative pronoun rule that coordinates the first occurrence of \texttt{\underline{ALICE}}:
\[
\input{reforderex2a.tikz}  
\]
Unless the initial sentence is simple, we cannot first use a subject relative pronoun rule followed by the reflexive pronoun rule, which would yield the \bR \textbf{ungrammatical sentence}\e:
\[
\bR \input{reforderex2b.tikz} \e
\]  
\end{example}

As a final remark, we note that by reversing the examples in this subsection, sentences involving relative pronouns can be broken down into lists of simple sentences.

\subsection{Compound sentence structure II: phrase scope}\label{sec:phrscope} 

Another way compound sentences arise is by sentences that contain subphrases that are themselves sentences. We introduce two such cases. We first introduce these cases intuitively, then we introduce the refinement of \emph{phrase scope} to eliminate a source of ambiguity.

\paragraph{Verbs with Sentential Complement.} We treat verbs with a sentential complement -- such as to \texttt{\underline{SEE}} or \texttt{\underline{THINK}} -- as their own grammatical class of verb. A sentence may consist of a noun-phrase, followed by a verb with sentential complement, followed by a sentence (e.g.~from \texttt{\underline{BOB DANCES.}} to \texttt{\underline{ALICE SEES BOB DANCE.}}).
\[
\begin{tabular}{|c|c|}
\hline
\text{Rule} & \text{Planar Tree-Fragment}  \\ \hline
$\texttt{S} \mapsto \texttt{NP} \cdot \texttt{SCV} \cdot \texttt{S}$ & \input{SCVsimple.tikz} \\ \hline
\end{tabular}
\]

\begin{example} For the example given above we obtain:
\[
\input{AsBdcsg.tikz}
\]
\end{example}

\paragraph{Conjunctions.}  Conjunctions we treat similarly to verbs with a sentential complement, except with two phrase-bounded regions rather than one, on each side of a conjunction (e.g.~~ from \texttt{\underline{BOB DANCES.}} and \texttt{\underline{ALICE LAUGHS.}} to  \texttt{\underline{BOB DANCES SO ALICE LAUGHS.}})  

\[
\begin{tabular}{|c|c|}
\hline
\text{Rule} & \text{Planar Tree-Fragment}  \\ \hline
$\texttt{S} \mapsto \texttt{S} \cdot \texttt{CNJ} \cdot \texttt{S}$ & \input{CNJsimple.tikz} \\ \hline
\end{tabular}
\]

\begin{example} We have:
\[
\input{AduBgHcsg.tikz}
\]
\end{example}

There is still a source of ambiguity to be addressed in the above formulation, which is where phrase scope comes into play.  For the sentence \texttt{\underline{ALICE SEES DRUNK BOB DANCE.}} there is only one phrase structure:
\[
\input{AsdBdcsg.tikz}
\]
which places all of \texttt{DRUNK BOB DANCE(S)} under the scope of what \texttt{ALICE SEES}. However, we wish to further distinguish between the following cases, where:
\begin{itemize}
    \item Alice sees \textbf{both} that Bob is drunk and that Bob dances.
    \item Alice sees \textbf{only} that Bob dances, but not that Bob is drunk.
\end{itemize}
To make the distinction between these two cases we introduce two  `formal' types \bB(\e and \bB)\e, which represent the left and right boundaries of phrase scope respectively. We re-express the rules for verbs with sentential complement and conjunctions to additionally express phrase boundaries:
\[
\begin{tabular}{|c|c|}
\hline
\text{Rule} & \text{Planar Tree-Fragment}  \\ \hline
$\texttt{S} \mapsto \texttt{NP} \cdot \texttt{SCV} \cdot \bB ( \e \ \cdot \ \texttt{S} \ \cdot \ \bB ) \e$ & \input{SCVscopecsg.tikz} \\ \hline
$\texttt{S} \mapsto \bB ( \e \ \cdot \ \texttt{S} \ \cdot \ \bB ) \e \cdot \texttt{CNJ} \cdot \bB ( \e \ \cdot \ \texttt{S} \ \cdot \ \bB ) \e$ & \input{CNJscopecsg.tikz} \\ \hline
\end{tabular}
\]
To model the permission of nouns to partially live outside the scope of a phrase as in the example above, we include the following rules that allow a \texttt{NP} to `cross the border' of a phrase boundary:\footnote{\label{footnote:cfarg2} 
	We may also recast these rules in a context-free manner. Rather than have context-sensitive scope changing rules, we may instead use marked variables $\texttt{X}^{n}$ to indicate that they are in scope, where the superscript can be a natural number to indicate scope depth. To spell this out as an example, we may replace the Sent.Comp.Verb rule
	$\texttt{S} \mapsto \texttt{NP} \cdot \texttt{SCV} \cdot \texttt{(} \cdot \texttt{S} \cdot \texttt{)}$
	with
	$\texttt{S}^{0} \mapsto \texttt{NP}^{0} \cdot \texttt{SCV}^{0} \cdot \texttt{S}^{1}$.
	Or, more generally
	$\forall k_{\in \mathbb{N}}: \texttt{S}^{k} \mapsto \texttt{NP}^{k} \cdot \texttt{SCV}^{k} \cdot \texttt{S}^{(k+1)}$.
	We may stay strictly context-free by bounding the index by a large finite number. So we may have families of context-free rules that are the same as before apart from matching scope indices, which, alongside the planarity of the graphs generated by the rules, ensures that all rules are applied within the same scope. For example, the rule for introducing an intransitive verb may be written
	$\forall k_{\in \mathbb{N}}: \texttt{S}^{k} \mapsto \texttt{NP}^{k} \cdot \texttt{IVP}^{k}$.
	In this setting, we may express the phrase-scope rules -- which just say that a noun may leave a scope -- as
	$\forall k_{>0}: \texttt{NP}^{k} \mapsto \texttt{NP}^{(k-1)}$.
	}
\[
\begin{tabular}{|c|c|}
\hline
\text{Rule} & \text{Planar Tree-Fragment}  \\ \hline
$\bB ( \e \ \cdot \ \texttt{NP} \mapsto \texttt{NP} \cdot \bB ( \e$ & \input{scopeexitcsg.tikz} \\ \hline
$\texttt{NP} \ \cdot \ \bB ) \e \mapsto \ \bB ) \e \ \cdot \texttt{NP}$ & \input{scopeexitRcsg.tikz} \\ \hline
\end{tabular}
\]

\begin{example}
To disambiguate that in the sentence
\[\texttt{\underline{ALICE SEES DRUNK BOB DANCE.}}\]
Alice sees \textbf{both} that Bob is drunk and that Bob dances, we have the following phrase structure:
\[
\input{AsdBdcsg1.tikz}
\]
To disambiguate that Alice \textbf{only} sees that Bob dances, and not that he is drunk, we have:
\[
\input{AsdBdcsg2.tikz}
\]
\end{example}
 
\begin{example}
In Figure \ref{fig:comic1} we illustrate how one can generate text with our hybrid grammar. 
\begin{figure}[h!]
    \centering
    \scalebox{0.75}{\input{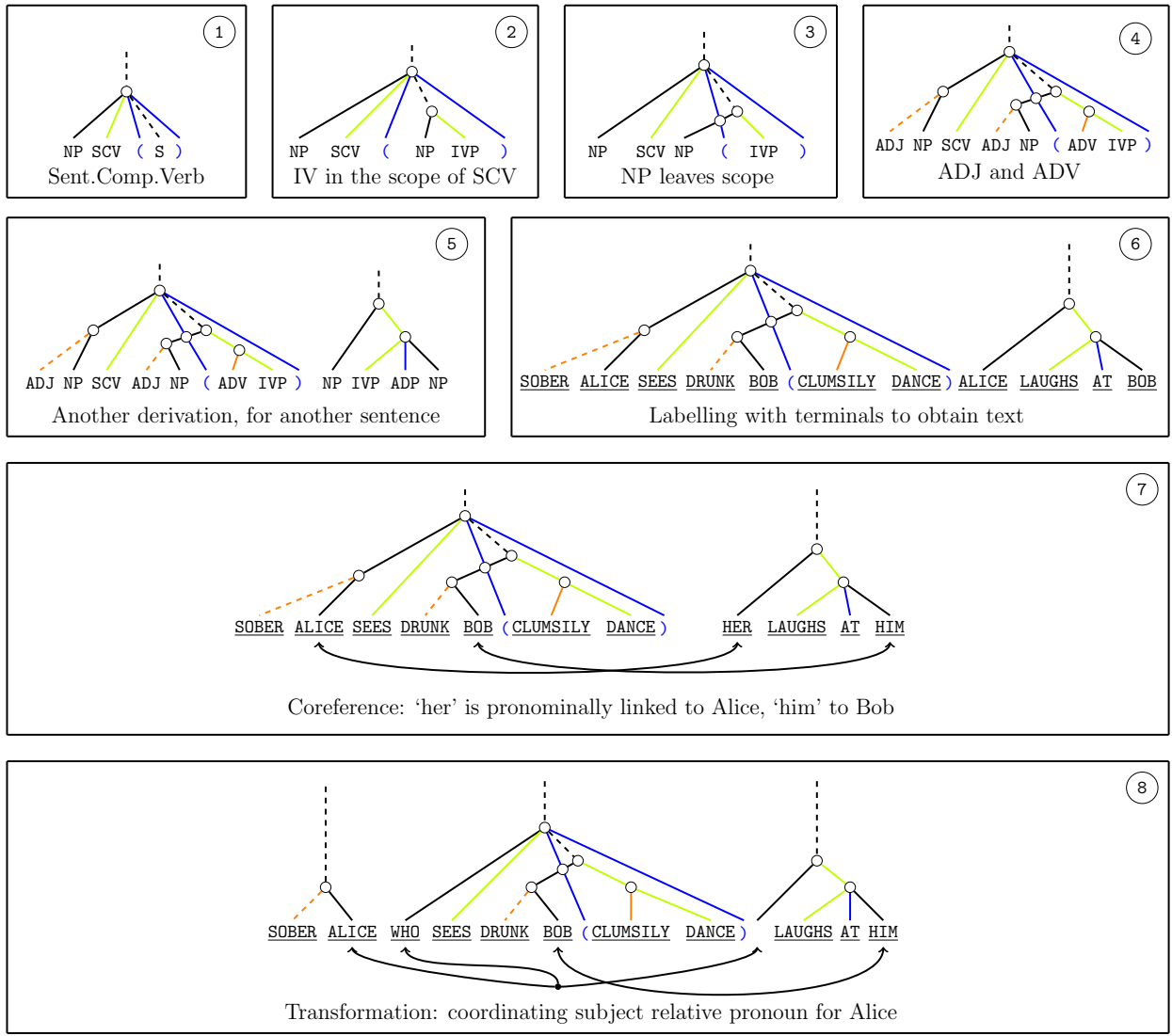}}
    \caption{Generating text with hybrid grammar, illustrated}
    \label{fig:comic1}
\end{figure} 
\end{example}

Going forward, we refer to our hybrid grammar of this section as ``hybrid grammar" or just ``grammar" when there is no confusion. 
 
\section{Text diagrams}\label{sec:congraphas}

We now introduce \emph{text diagrams}, which incorporate the grammatical phenomena of text with hybrid grammar (cf.~pronominal links and phrase boundaries) into a single mathematical structure. We will do so by means of a systematic passage from text with hybrid grammar to text diagrams. Artefacts required to handle the behaviour of relative pronouns in hybrid grammar vanish in the setting of text diagrams.

\subsection{Preliminaries}

String diagrams \cite{Penrose, JS, SelingerSurvey, CKbook} are a graphical mathematical framework for composing input-output boxes.  For us the `boxes' will have the following shape:
\[
\input{stringscopy.tikz}
\]
For composition one typically one distinguishes between parallel composition:
\[
\input{parallelstrings.tikz}
\]
or sequential composition by plugging together matching outputs and inputs. 
\[
\input{opengraph_compex.tikz} 
\]
Despite the 1-dimensionality connoted by `string', string diagrams in general also accommodate higher-dimensional structures such as planes and volumes \cite{reutter_high-level_2019}.



\subsection{Simple sentences as text diagrams}\label{gramtograph}

The initial symbol for our phrase structure rules in the previous section was \texttt{S}.  In the passage from hybrid grammar to text diagrams, systematically: 
\bit
\item We will replace the sentence type \texttt{S} with a \underline{sentence-dependent} number of \texttt{NP} wires. 
\eit
In this way, tree-shaped texts with hybrid grammar become string diagrams with one input \texttt{NP} wire for every pronominally distinct \texttt{NP} label in the text. This geometric change of perspective is what enables a passage to text circuits. To achieve this, we alter the rules of Section \ref{sec:grammar} as follows:
\begin{itemize}
    \item We remove \texttt{S} types.
    \item Every rule must preserve the number of input and output noun wires, so phrase structure rules that introduce \texttt{NP} types must be altered to also take the same number of \texttt{NP} types as input.
\end{itemize}
In the table below where we depict just those rules that are altered in this way. As can be seen, besides removing sentence-inputs and adding  noun-inputs, subscripting \texttt{NP} where there are multiple, we also vertically align matching input-output pairs:
\[
\begin{tabular}{|c|c|c|}
\hline
\text{Grammar} & \text{Rule} & \text{Diagram}  \\ \hline
\input{IVcsgcopy.tikz} & \text{Intrans.Verb} & \input{IVgraphcopy.tikz} \\ \hline
\input{TVcsgcopy.tikz} & \text{Trans.Verb} & \input{TVgraphcopy.tikz} \\ \hline
\input{ADJiscsgcopy.tikz} & \text{Adjective} & \input{ADJisgraphcopy.tikz} \\ \hline
\input{ADPIVcsgcopy.tikz} & \text{Adposition(IV)} & \input{ADPIVgraphcopy.tikz} \\ \hline
\input{ADPTVcsgcopy.tikz} & \text{Adposition(TV)} & \input{ADPTVgraphcopy.tikz} \\ \hline
\end{tabular}
\]
Similarly, when we draw text diagrams, as a visual convention  we consistently arrange the diagram so that matching input and output \texttt{NP} wires align, and bend those a bit always pointing either upward or downward, to better reflect the notion that wires should be considered as inputs and outputs in the string diagram sense: 
\[
\input{diagramconvention.tikz}
\]
We can now also drop one of the doubly-appearing labels:
\[
\input{diagramconvention2.tikz}
\]

\begin{example} 
For the simple sentence \texttt{\underline{ALICE GIVES BEER TO BOB}}, the grammar is:
\[
\input{AgBtB2.tikz}
\]
from which we obtain the following text diagram:
\[
\input{AgBtBgraph.tikz}
\]
\end{example}

\begin{example}
We allow the new input \texttt{NP} wires obtained from grammar to diagram translation to cross other wires to reach the top of the diagram. For eample, in the sentence \texttt{\underline{ALICE GIVES BEER TO BOB FOR CLAIRE}} the grammar is:
\[\input{AgBtBfC.tikz}\]
from which we obtain the following text diagram, where noun wires are vertically aligned:
\[\input{AgBtBfCdiagram.tikz}\]
\end{example}

\subsection{Rewriting text diagrams}

When dealing with relative pronouns we have seen something analogous to Chomskian phrase-structure transformations: operations that fuse and more generally transform text  diagrams. We formalise these as \underline{diagram rewrites}, each of which replaces a diagram with another that has the same input and output wires as the first. We illustrate examples of such rewrites in the sections to follow.

\subsection{Pronominal links in text diagrams}\label{sec:prondiag}  

Pronominal links take a different shape in text diagrams than in grammar. 
The pronominal link arrows of \ref{sec:pronom-links} become wires that link output and input noun wires in a chain\footnote{We could have also had the pronominal links in the chain go from northwest to southeast. We choose this direction such that the gates of the resulting text circuit, read top-to-bottom and left-to-right, reflect the order in which words appear in text.}:
\[
\input{link2diagram.tikz}
\]
where the following pieces relate pronominal link wires to noun wires:
\[
\input{plinkgens.tikz}
\]
These pieces can vanish by the following link-elimination rewrites:
\[
\input{prontransforms.tikz}  
\]
In particular, this means that where $\mathcal{D}_1$ and $\mathcal{D}_2$ are text diagrams, we have:
\[
\input{pronlinkrule.tikz}  
\]  
\[
\input{pronlinkrule2.tikz}
\]
As we have seen now, wires may cross one another: this is an important quality of text diagrams.
\begin{convention}[In text diagrams, only connectivity matters]\label{conv:onlyconnect}
While hybrid grammar diagrams are planar, in text diagrams wires and labels may be freely induced to cross over one another, so long as input-output connectivity is preserved.
\end{convention}
Link-elimination rewrites may induce twists in noun wires; a consequence of an overarching principle that only output-to-input connectivity matters. So we also have the following link-elimination rewrites:
\[
\input{prontransforms2.tikz}  
\]
We demonstrate by means of examples how these rewrites interpret pronominal links as composition of text diagrams.  

\begin{example}[Subject relative pronouns revisited]
We start with the following text:
\[
\input{srpex1.tikz}
\]
For which we can replace each tree with a text diagram:
\[
\input{srpexA2.tikz}
\]
Now we can replace the pronominal link by its text diagram counterpart. Now \texttt{BOB} (which is a proper noun, rather than the pronoun \texttt{WHO}) labels the only available output wire:
\[
\input{srpexA3.tikz}
\]
The pronominal link can now be rewritten away:
\[
\input{srpexA4.tikz}
\]
\end{example}

\begin{example}[Subject relative pronouns revisited]
Pronominal links in the setting of text diagrams `explain' the artefacts that were necessary in hybrid grammar to accommodate relative pronouns. Recall that for the special rule for subject relative pronouns, we obtain a structure with \texttt{!} and \textvisiblespace \ artefacts:
\[
\input{ssrpex.tikz}
\]
When we move to text diagrams, the \texttt{!} artefact disappears:
\[
\input{ssrpexA2.tikz}
\]
Now we replace the pronominal links with their text diagram counterparts, obtaining:
\[
\input{ssrpexA3.tikz}
\]
Rewriting away pronominal links, we have the following text diagram:
\[
\input{ssrpexAF.tikz}
\]

\end{example}



\subsection{Phrase scope as phrase bubbles.}  

We will have to deal with compound sentences differently, because in the passage from text with grammar to text diagrams, we have replaced the sentence type \texttt{S} by a sentence-dependent collection of noun wires. This requires the use of string diagrams with regions.\footnote{We forgo exposition here, but everything we present here can be expressed in the setting of associative $n$-categories \cite{dorn_associative_2018}, and in fact prototyped and implemented in a proof assistant \cite{reutter_high-level_2019}.} We will introduce  generators of text diagrams with regions that identify the boundaries that delineate a sentential subphrase. For verbs with a sentential complement that take a phrase to the right, we have diagram pieces that allow a noun-wire to enter a bounded phrase from the right, and exit on the left:
\[
\input{phrenter.tikz} \quad\quad\quad\quad \input{phrexit.tikz}
\]
For conjunctions, which also take a phrase to the left, we have rewrites that allow a noun-wire to enter a bounded phrase from the left, and exit on the right:
\[
\input{phrenterR.tikz} \quad\quad\quad\quad \input{phrexitR.tikz}
\]
In both cases, we have one more rewrite to `cap off' phrase regions. Formally we must distinguish between phrases-to-the-left and -to-the-right, which have differing caps, but in the graphical presentation we elide this distinction as it is visually evident:
\[
\input{phrcap.tikz}
\]

\begin{convention}[Rules for phrase scope]\label{conv:phrscope}
We state graphical conventions for phrase scope and phrase regions here. First, phrase scoped regions behave as planar obstacles in a text diagram, except for \texttt{NP} wires and pronominal links. This has two consequences: that nodes and wires occurring within a phrase scope are `trapped inside', and also that multiple phrase scope constructions may nest iteratively, but no two overlap. Second, \texttt{\underline{NP}} labels may only occur outside phrase scope.
\end{convention}

For phrase scope constructions in grammar -- verbs with sentential complements and conjunctions -- in order to ensure that the resulting text diagram has an equal number of input and output noun wires, our correspondence must take into account the number of labelled noun wires (drawn with a `foot') in the subsentence within the phrase scope. We subscript the noun labels with indices $i$ and $o$, for `inside scope' and `outside scope'. We have diagram counterparts for verbs with sentential complements and conjunctions as follows: 
\[
\begin{tabular}{|c|c|c|}
\hline
\text{Grammar} & \text{Rule} & \text{Diagram}  \\ \hline
\input{SCVhybrid.tikz} &  \text{Sent.Comp.Verb} &\input{SCVtranslate.tikz} \\ \hline
\input{CNJhybrid.tikz} & \text{Conjunction} &  \input{CNJtranslate.tikz}   \\ \hline
\end{tabular}
\]

\begin{example}  
For \texttt{\underline{ALICE SEES BOB DANCE}} we now obtain:
\[
\input{AsBdgraph.tikz}
\]
In this example, the sentential subphrase \texttt{BOB DANCE(S)} is `contained' in a grey `phrase bubble'. Recalling Convention \ref{conv:phrscope}, although text diagrams in general are not restricted by planarity, we impose the condition that -- apart from noun wires which have specific diagrams that allow them to -- no wires penetrate the boundary of the `bubble'; in this way, the phrase regions capture the behaviour of phrase scope viewed as subtrees, or more generally as in dependency grammars.
\end{example}


\section{Text circuits}\label{sec:circs}  


\subsection{Definition}

Our \emph{text circuits} are made up of three ingredients:
\bit
\item wires
\item boxes, or gates
\item boxes with holes that fit a box, or 2nd order gates
\eit
Firstly, nouns are represented by wires, each `distinct' noun having its own wire:
\[
\input{nounwiresABN.tikz} 
\]
We represent adjectives, intransitive verbs, and transitive verbs by gates acting on noun-wires: 
\[
\input{ADJgate.tikz} \quad\quad\quad \input{IVgate.tikz} \quad\quad\quad \input{TVgate.tikz}
\]
Since a transitive verb has both a subject and an object  noun, that will then be two noun-wires, while adjectives and intransitive verbs only have one. 

Adverbs, which modify verbs, we represent as boxes with holes in them, with a number of dangling wires in the hole indicating the shape of gate expected, and these should match the input- and output-wires  of the box with the whole:
\[
\input{ADVbox.tikz}
\]
Similarly, adpositions also modify verbs, by moreover adding another noun-wire to the right:
\[
\input{ADPIVbox.tikz}
\]
For verbs that take sentential complements and conjunctions, we have families of boxes to accommodate input circuits of all sizes. They add another noun-wire to the left of a circuit:\footnote{The `dot dot dot' notation within boxes is graphically formal \cite{wilson_string_2022}. Interpretations of such boxes were earlier formalised in \cite{merry_reasoning_2014,quick_-logic_2015,zamdzhiev_rewriting_2017}.}
\[
\input{SCVbox.tikz}
\]
while conjunctions are boxes that take two circuits which might share labels on some wires:
\[
\input{CNJbox2.tikz}
\]
As special cases, when the noun-wires of two circuits are disjoint (left), or coincide (right), conjunctions are depicted as follows:
\[
\input{CNJbox.tikz} \qquad\qquad\qquad\qquad \input{CNJbox3.tikz}
\]

Of course filled up boxes are just gates
\[
\input{ADPIVgate.tikz}
\]
and we will now discuss how those compose. Gates compose sequentially by matching labels on some of their noun-wires and in parallel when they share no noun-wires,  to give \underline{text circuits}, which by convention we read from top-to-bottom:  
\[
\input{gatecompex1.tikz}  
\]

\begin{convention}\label{conv:sliding}
Sometimes we allow wires to twist past each other, and we consider two circuits the same if their gate-connectivity is the same:
\[
\input{gatecompex1.tikz} \quad\quad = \quad\quad \input{gatecompex2.tikz}
\]
Since only gate-connectivity matters, we consider circuits the same if all that differs is the horizontal positioning of gates composed in parallel:
\[
\input{gateeqslide.tikz} 
\]
We do care about output-to-input connectivity, so in particular, we {\bR\bf do not\e}
consider circuits to be equal up to sequentially composed gates commuting past each other:  
\[
\input{gateneqcommute.tikz}  
\]
\end{convention}

\begin{example}  
The sentence \texttt{\underline{ALICE SEES BOB LIKES FLOWERS THAT CLAIRE PICKS}} can intuitively be given the following text circuit:
\[
\input{multigateholeex.tikz}
\]
\end{example}

\begin{example}
Figure \ref{fig:circuitgen} illustrates how to directly generate text circuits, by providing an example analogous to the text generated in Figure \ref{fig:comic1} where we used our hybrid grammar.
\begin{figure}[h!]
    \centering
    \scalebox{1}{\input{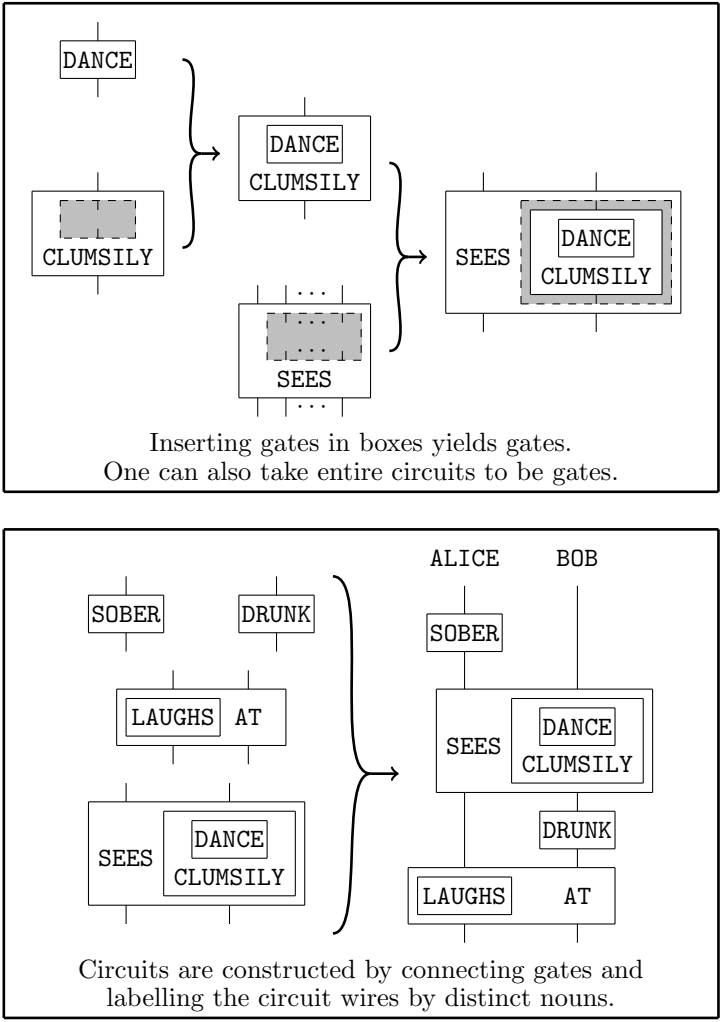}}
    \caption{Generating text circuits directly.}
    \label{fig:circuitgen}
\end{figure}
\end{example}

\section{Mathematical results}\label{sec:thm}

\subsection{Main Text Circuit Theorem}\label{thm:main}
We assume some finite vocabulary of words with grammatical types we have considered so far.
Let $\mathfrak{T}$ denote the set of all raw text generated by our grammar.
Let $\mathfrak{C}$ denote the set of all text circuits. We prove the following results:
\begin{enumerate}
    \item We show that the translation rules of Section \ref{graph2gateredux} algorithmically associate any  text with  grammar to a text circuit.
    \item All text circuits are obtainable in this way.
\end{enumerate}

\begin{theorem}
The translation rules of Section \ref{graph2gateredux} define a surjection $\mathfrak{T} \twoheadrightarrow \mathfrak{C}$. 
\end{theorem}

\subsection{Refinements, extensions, conventions}\label{sec:asscon}

There are clarifications to be made before proving the Theorem, which is a claim about \emph{all} text circuits, and \emph{all} raw texts generated by hybrid grammar. Convention \ref{conv:wireorder} clarifies what exactly counts as a text circuit for our purposes. We address an edge-case circuit -- where a wire appears in a box unconnected to any gates -- in Refinement \ref{asm:exists}. For the proof-strategy of Lemma \ref{lem:surj}, which relies on a step that treats multiple gates in parallel as single sentences, we accommodate textual elements such as commas and `contentless' conjunctions such as \texttt{AND ALSO} by means of a generic conjunction \texttt{[\&]}. We note in Refinement \ref{conv:complement} that the phrase-scope distinctions of hybrid grammar are properly expressible in raw text by means of a generic contentless complementiser \texttt{[THAT]}. Finally, we extend the definition of text circuits to include `reflexive pronoun boxes' as structural components.

\begin{convention}[Wire ordering]\label{conv:wireorder}
We elaborate Convention \ref{conv:sliding} for depicting circuits. Every wire of a circuit is labelled with a noun, and when depicted these noun-labels are forced to take a left-to-right order. By convention, this order is kept the same for inputs and outputs of a circuit, by introducing wire twists where necessary. Note that two circuits with the same connectivity but different orders of noun-labels are still considered equal:
\[
\input{wiretwisteq.tikz}
\]
To give a negative example, the following diagram \textbf{\bR violates\e} the visual convention, because the input and output wires have a different noun order:
\[
\bR\input{twistviolation.tikz}\e
\]
The contents of a box are  circuits as well, which must obey the same input-output agreement rules:
\[
\input{validboxex.tikz} \quad\quad\quad\quad\quad\quad \bR \input{invalidboxex.tikz} \e
\]
\end{convention} 

\begin{refinement}[The verb \texttt{EXISTS}]\label{asm:exists}
We introduce a special structural rule for an identity wire that does not participate in any gates. In these cases, we interpret the identity wire as the special transitive verb \texttt{EXISTS} at the level of raw text. The rewrite rule for \texttt{EXISTS} from text diagrams to circuits is as follows:
\[
\input{existsfig.tikz}
\]
\end{refinement}  

\begin{refinement}[Sentence composition using \texttt{[\&]}]\label{asm:conj}
So far, we have mostly considered listing sentences as text to compose them, e.g.:
\[
\texttt{\underline{ALICE RUNS. BOB DRINKS.}}
\]
For the purposes of Lemma \ref{lem:surj}, we define a special conjunction \texttt{[\&]} at the level of text, which allows us to consider multiple gates in parallel as arising from a single sentence in an unambiguous way. In English, there are multiple `contentless' ways that two sentences may be composed as a single sentence. For example, by an ampersand:
\[
\texttt{\underline{ALICE RUNS [\&] BOB DRINKS.}}
\]
or by `contentless' conjunctive phrases such as \texttt{AND ALSO}~e.g.: 
\[
\texttt{\underline{ALICE RUNS AND ALSO BOB DRINKS.}}
\]
The special conjunction \texttt{[\&]} is intended to be a catchall for these kinds of `contentless' conjunctions. At the level of circuits \texttt{[\&]} is interpreted as gate composition, and its rewrite rule from text diagrams to circuits is as follows:
\[
\input{sentconj.tikz}
\]
Note that in the case where the noun wires are disjoint, the rewrite rule for \texttt{[\&]} just yields parallel composition of circuits:
\[
\input{sentconjdisj.tikz}
\]
\end{refinement}

\begin{refinement}[Sentence composition within phrase scope, and the complementizer \texttt{[THAT]}]\label{conv:complement}
Consider the sentence:
\[
\texttt{\underline{CLAIRE SEES ALICE RUNS \bR[\&] BOB DRINKS\e.}}
\]
For us, when the raw text is equipped with hybrid grammar structure, there is no ambiguity as to whether Claire sees both that Alice runs and Bob drinks, or just the former. Recalling Refinement \ref{asm:conj}, neither reading of \texttt{[\&]} as a comma or \texttt{AND ALSO} resolves this ambiguity. Since our broad claim is that we address a restriction of natural language, we have to justify that this distinction made by hybrid grammar structure is one that actually has a counterpart at the level of raw text. To disambiguate in a similar manner as our previous refinements, we introduce a complementiser \texttt{[THAT]}, which behaves much like the phrase scope formal types \texttt{(} and \texttt{)} introduced in Section \ref{sec:phrscope}. Complementisers are words such as \texttt{HOW} or \texttt{WHAT}, which prefix sentential complements~e.g.:
\[
\texttt{\underline{CLAIRE SEES HOW ALICE RUNS.}}
\]
\[
\texttt{\underline{CLAIRE SEES WHAT BOB DRINKS.}}
\]
We do not consider complementisers in general here. Instead we use just one `contentless' complementiser \texttt{[THAT]}, which we freely omit when it is implicit. Returning to the initial example, the presence of a complementiser allows us to distinguish, at the level of raw text, the case where Claire \textbf{only} sees Alice running:
\[
\texttt{\underline{CLAIRE SEES [THAT] ALICE RUNS [\&] BOB DRINKS.}}
\]
from the case where Claire sees \textbf{both} that Alice runs and that Bob drinks:
\[
\texttt{\underline{CLAIRE SEES [THAT] ALICE RUNS [\&] [THAT] BOB DRINKS.}}
\]
So, when we encounter the conjunction \texttt{[\&]} within a phrase scope, we interpret it at the level of text as \texttt{AND ALSO THAT}. More generally, we use \texttt{[THAT]} to signify the open-bracket that begins a scoped phrase in raw text.
\end{refinement}

%

\begin{convention}[Reflexive pronoun boxes]\label{asm:reflpron}
We extend the definition of text circuits to accommodate reflexive pronouns, such as \texttt{HERSELF}, \texttt{ITSELF}. We treat pronominal links as gate composition in all other cases, but by definition, a reflexive pronoun in a circuit occurs when an output noun wire must compose with an input noun wire in its relative past, which cannot be drawn as a circuit. So, at the level of circuits, we assert a family of `reflexive pronoun boxes'. For all $k > 1 \in \mathbb{N}$, and for all pairs of indices $1 \leq i < j \leq k$, we have a box:
\[
\input{reflpronbox.tikz}
\]
These boxes are the targets of rewrites from text diagrams when a pronominal link elimination would connect an output of a circuit to an input, `doubling back' a noun wire:
\[
\input{reflboxredux.tikz} 
\]
\end{convention}

\begin{refinement}[Coherence of reflexive pronoun boxes]\label{ref:proncoherence} 
A \emph{critical pair} in a rewrite system is a situation in which two rules are applicable to a term, and result in different final outcomes. Critical pairs are hence the only obstacle to the claim of Proposition \ref{lem:function}. Reflexive pronouns are involved in all critical pairs that arise in rewriting text diagrams into circuits. Consider the following example, for the text diagram corresponding to the text:
\[\texttt{\underline{BOB DRINKS. BOB LIKES BOB.}}\]
\[\input{criticalpair.tikz}\]
So the order of pronominal link eliminations and (their special case) reflexive pronoun box introductions matters. As we show in Corollary \ref{cor:reflnorm}, we may generally deal with these critical pairs by a convention that all reflexive pronoun box introductions come before the elimination of other pronominal links. Recalling the footnote in Section \ref{sec:pronom-links}, a pronominal link in a text diagram can take two directions. Neither is strictly appropriate in the case of reflexive pronouns, where a pronominal link should \emph{identify} two wires. So here we introduce syntactic sugar for the reflexive pronoun box, and introduce rewrite rules such that identification of wires is respected. We introduce the following shorthand for reflexive pronominal links:
\[\input{syntacticsugar1.tikz}\]
Identifying wires is \emph{associative}, which we enforce by the following bidirectional rewrites:
\[\input{reflfuse.tikz}\]
A reflexive pronoun box with only identity wires inside can be eliminated by an \emph{identity} rule:
\[\input{reflid.tikz}\]
Conversely, we can introduce new reflexive pronoun boxes by \emph{splitting} existing ones along identified wires:
\[\input{reflid2.tikz}\]
And finally, a special aspect of reflexive pronoun boxes is that gates on one wire may \emph{slide} out of them. This reflects the fact that sentences that only modify a single noun never require a reflexive pronoun, which would refer to another noun argument:
\[\input{reflsingleslide.tikz}\]
\end{refinement}

\subsection{Text diagrams to text circuits}\label{graph2gateredux}

This section presents lemmas that together constitute the hard work in the proof of Proposition \ref{lem:function}, which claims the existence of a function $\mathfrak{T} \rightarrow \mathfrak{C}$.

First, Lemma \ref{lem:shrinking} and Corollary \ref{cor:reflnorm} constructively organise the rewrites of Refinement \ref{ref:proncoherence} into a function from text diagrams to text diagrams with no pronominal links.
Second, Lemma \ref{lem:gatenorm} introduces new rewrite rules and ancilla types which constructively constitute a function that takes text diagrams with no pronominal links nor phrase scoping into a normal form that corresponds to a single circuit gate.
Third, Lemma \ref{lem:gatenorm} is extended by Lemma \ref{lem:diagnorm} to account for text diagrams with phrase scope as well.
When taken together with the content of Section \ref{sec:congraphas} -- which provides a correspondence that sends every hybrid grammar text to a text diagram -- the three stages above complete the description of a function from hybrid grammar text to text circuits.

\begin{lemma}[Shrinking reflexive pronouns]
\label{lem:shrinking}
The associativity, identity, splitting, and sliding rewrite rules for reflexive pronouns suffice to recast text circuits in a form where every reflexive pronoun box contains exactly one gate.
\begin{proof}
Nested reflexive pronoun boxes can be fused and rearranged according to the associativity rule. Any reflexive pronoun box containing two or more gates composed sequentially along an identified wire can be split by the splitting rule. The splitting and identity rules may be applied to eliminate twists in reflexive pronoun boxes containing a single gate, such that the inputs and outputs align.
\end{proof}
\end{lemma}

\begin{corollary}\label{cor:reflnorm}
Text circuits obtained by the shrinking lemma coincide with text circuits obtained from text diagrams where reflexive pronoun box reductions are applied before other pronominal link eliminations.
\begin{proof}
By construction, reflexive pronouns only occur within a sentence, and sentences correspond to individual gates. The rewrite rules of Lemma \ref{lem:shrinking} preserve gate connectivity, which is determined by non-reflexive pronominal links.
\end{proof}
\end{corollary}

\begin{example}[An application of the shrinking lemma]
Consider a text circuit that glosses as the text:
\[\underline{\texttt{BOB TELLS BOB ABOUT BOB. BOB DRINKS. BOB LIKES BOB.}}\]
\[\input{reflreductionexample.tikz}\]
\end{example}

\begin{figure}[h!]
\centering
\begin{tabular}{|c|c|}
\hline
Rule name & Text diagram rewrite \\ \hline
\texttt{IS}-elimination & \input{ADJistransform.tikz} \\ \hline
\texttt{ADV}-gather & \input{advgather.tikz} \\ \hline
\texttt{ADV}-assoc. & \input{advassoc.tikz} \\ \hline
\texttt{ADP(IV)}-ancilla & \input{adpIVancilla.tikz} \\ \hline
\texttt{ADP(TV)}-ancilla & \input{adpTVancilla.tikz} \\ \hline
\texttt{adp}-gather & \input{adpgather.tikz}  \\ \hline
\texttt{adp}-assoc. & \input{adpassoc.tikz} \\ \hline
\texttt{adp-ADV}-order & \input{adorder.tikz} \\ \hline
\end{tabular}
\caption{Gate normalisation rewrites}
\label{gatenormalisationrules}
\end{figure}

\begin{lemma}\label{lem:gatenorm}
We present ancillas and rewrites in Table \ref{gatenormalisationrules} such that every text diagram without pronominal links and without phrase scope can be viewed as a unique text circuit constructed out of gates of the following forms:
\[\input{gatenormals.tikz}\]
\begin{proof}

We introduce an ancillary wire type \texttt{adp}, along with the following ancillary text diagram components.

\[\input{ancillas.tikz}\]

We present normalisation rewrite rules in the table of Figure \ref{gatenormalisationrules}.

The \texttt{NP} wires of a text diagram correspond to the noun wires in a text circuit. For a text diagram from a hybrid grammar structure without pronominal links and phrase scope, there are six non-label text diagram nodes that involve \texttt{NP} wires; each such node corresponds to either a gate or a box.
\[(1) \input{ADJcsg.tikz} (2) \input{ADJisgraph.tikz} (3) \input{IVgraph.tikz} (4) \input{TVgraph.tikz} (5) \input{ADPIVgraph.tikz} (6) \input{ADPTVgraph.tikz}\]
The \texttt{IS}-elimination rule recasts all instances of case (2) nodes as case (1) nodes. Recalling by Convention \ref{conv:onlyconnect} that in text diagrams only connectivity matters, we may contract the \texttt{\underline{ADJ}} label associated with each instance of case (1), interpreting as a gate in a text circuit as follows.
\[\input{adjnormalex.tikz}\]
Cases (5) and (6) are adpositions, which can only appear in the presence of an \texttt{IV} or \texttt{TV} wire respectively, which we refer to collectively as a \texttt{V} wire for simplicity. The only way such wires appear is by cases (3) and (4) respectively, and the only way they end is by \texttt{\underline{V}} labels.

We cannot repeat the same contraction trick as in \texttt{ADJ} labels for \texttt{V} labels directly, because in between a node of case (3) or (4) and its associated \texttt{\underline{V}} label, there may be \texttt{ADV} nodes, or \texttt{ADP} nodes as in (5) and (6). However, all such nodes have one input and one output \texttt{V} wire, so all \texttt{ADV} and \texttt{ADP} nodes, and their labels, can be unambiguously associated with a single \texttt{\underline{V}} label.

The ancillas and rewrites are in service of obtaining a normal form for the \texttt{ADV} and \texttt{ADP} nodes and labels contracted to their parent \texttt{V} node of case (3) or (4). We analyse the three possible cases for a given \texttt{V} wire in the following order to illustrate the purpose of the rules; only \texttt{ADV} nodes appear; only \texttt{ADP} nodes appear; both kinds of nodes appear.

In the first case, when two \texttt{ADV} nodes appear adjacently on the same \texttt{V} wire, we apply the \texttt{ADV}-gather rule. The \texttt{ADV}-assoc bidirectional rewrite rule makes the order of \texttt{ADV}-gather applications irrelevant in the case of multiple adverbs for the same verb. Contracting labels, these rules allow re-expression of multiple \texttt{ADV} nodes on the same \texttt{V} wire as follows:
\[\input{advgatherex.tikz}\]
In the second case -- which also addresses nodes of type (5) and (6) -- when there are multiple \texttt{ADP} nodes, we first apply the \texttt{ADP}-ancilla rules to introduce the \texttt{adp} ancilla type. We may then apply \texttt{adp}-gather, mirroring the treatment for \texttt{ADV} wires: similarly, the \texttt{adp}-assoc rule makes the order of \texttt{adp}-gather rewrites irrelevant in the case of multiple adpositions on the same verb. These rules allow re-expression of multiple \texttt{ADP} nodes on the same wire as follows, which we illustrate for two \texttt{ADP(TV)}.
\[\input{adpgatherex.tikz}\]
In the third and final case where there are \texttt{ADV} and \texttt{ADP} nodes on a \texttt{V} wire, we first eliminate all \texttt{ADP} nodes of type (5) or (6) by \texttt{ADP}-ancilla rewrites. Then by applying the \texttt{adp}-\texttt{ADV}-order rule, we may arrange all \texttt{adp} nodes above all \texttt{ADV} nodes on the \texttt{V} wire, then we apply the treatment of the previous two cases of only \texttt{ADV} or only \texttt{adp} nodes.
Contracting \texttt{\underline{ADV}} and \texttt{\underline{ADP}} labels, this procedure obtains the following normal forms for every node of type (3) and (4) respectively.
\[\input{gatenormalformIV.tikz} \qquad\qquad\qquad \input{gatenormalformTV.tikz}\]
Which we may read as gates of the following form respectively, colour-coded for legibility.
\[\input{gatenormalIV.tikz} \quad\quad\quad\quad \input{gatenormalTV.tikz}\]
These are equivalent up to syntactic sugar to the last two gates in the claim. Uniqueness of the resulting circuit follows because none of the rules change the relative connectivity of nodes of type (1) through (4).
\end{proof}
\end{lemma}

\begin{lemma}\label{lem:diagnorm}
Every text diagram without pronominal links can be viewed as a unique text circuit.
\begin{proof}
We only need extend the claim of Lemma \ref{lem:gatenorm} to accommodate phrase scope. Recall by Convention \ref{conv:phrscope} that phrase scope only allows \texttt{NP} wires and pronominal links to pass through, that \texttt{\underline{NP}} labels may only occur outside of phrase scope, and that phrase scope nesting anywhere in the diagram is unambiguous. We prove the claim by (strong) induction, where the inductive hypothesis is that all text diagrams with phrase scope that nest at most $k$-levels deep can be viewed as a unique text circuit.
For the base case, where there is no phrase scope, we are done by Lemma \ref{lem:gatenorm}.
For induction, assume we have a text diagram with phrase scope nesting of depth $k+1$. The topmost phrase scope node is either an \texttt{SCV} node or a \texttt{CNJ} node.
\paragraph{Verbs with sentential complement} Since the same number of \texttt{NP} wires must enter as leave the phrase scope, by the inductive hypothesis we may express the content of phrase the phrase scope as a text circuit $\mathcal{C}$, which we may unambiguously read off as follows (modulo adverbs and adpositions for the \texttt{\underline{SCV}} label as considered previously.)
\[\input{SCVredux.tikz}\]
The remainder of the text diagram falls to the inductive hypothesis.
\paragraph{Conjunctions}
Similarly as before, we may express the contents of the left and right conjuncts as text circuits $\mathcal{C}_1$ and $\mathcal{C}_2$. There are two subcases we consider in order: where the conjuncts do not share \texttt{NP} wires, and when they do. In the former subcase, we obtain the following gate from a \texttt{CNJ} node.
\[\input{CNJredux.tikz}\]
In the latter subcase where the two conjuncts share noun wires -- or, anticipating later work, when there are pronominal links between the conjuncts -- we can introduce a conjugate box with overlapping arguments as syntactic sugar standing in for a `normal' \texttt{CNJ} box inside reflexive pronoun boxes:
\[
\input{CNJredux2.tikz}
\]
\end{proof}
\end{lemma}

\begin{example}
Figure \ref{fig:comic2} illustrates how to obtain text circuits from text diagrams.

\begin{figure}[!]
    \centering
    \scalebox{0.75}{\input{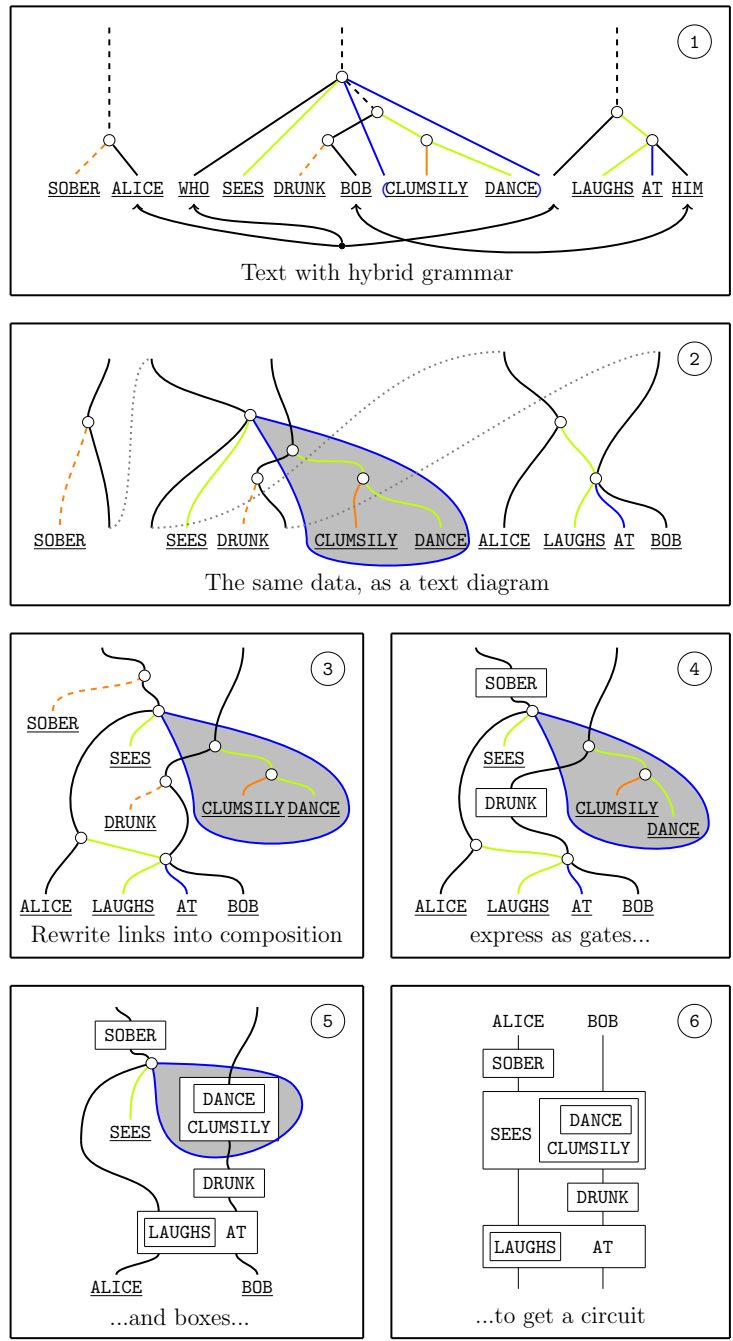}}
    \caption{Obtaining text circuits from text diagrams.} 
    \label{fig:comic2}
\end{figure}
\end{example}

\subsection{Proof of Theorem}

It suffices to demonstrate the following. First, that the translation procedure from text to circuit given in Section \ref{graph2gateredux} yields a function $\mathfrak{T} \rightarrow \mathfrak{C}$. Second, that this function is surjective.


\begin{proposition}\label{lem:function}  
 We have a function $\mathfrak{T} \rightarrow \mathfrak{C}$.
\begin{proof}
 Lemma \ref{lem:diagnorm} handles all text diagrams obtained from text without pronominal links. It remains to be shown that pronominal links can be resolved uniquely in the setting of text diagrams. The following transformations eliminate pronominal links.
\[
\input{prontransforms.tikz}  
\]
\[
\input{prontransforms2.tikz}  
\]
The only obstacle to interpreting these transformations directly as composition of text circuits is addressed by reflexive pronoun boxes, in Convention \ref{asm:reflpron}. The critical pairs that arise as a result of reflexive pronoun boxes are addressed by Refinement \ref{ref:proncoherence}, and a normal form for reflexive pronoun boxes is constructively provided by Lemma \ref{lem:shrinking}. So the initial hybrid text determines the corresponding circuit.
\end{proof}
\end{proposition}

\begin{proposition}\label{lem:surj}
$\mathfrak{T} \rightarrow \mathfrak{C}$ is surjective.
\begin{proof}
Surjectivity amounts to showing that an arbitrary circuit corresponds to a text that translates into that circuit; in other words, every text circuit can be textualised. First, we divide the circuit into horizontal slices, such that every slice either contains at least one gate, or contains only twisting wires. Here we exploit Refinement \ref{asm:exists}; when a wire inside the context of a box does not connect to any gates, we assume it connects to an  \texttt{EXISTS} gate. We iterate this slicing within the holes of boxes:  
\[
\input{circ2textex1.tikz} 
\]
Now we exploit Refinement \ref{asm:conj}; every horizontal slice with gates obtained this way is a collection of gates composed in parallel, possibly with identity wires. Identity wires inside the holes of boxes that do not connect to gates we treat as connected to an \texttt{EXISTS} gate, as in the hole of \texttt{SEES} in the example. Other identity wires we do not need to mention in text.  We deal with adjective gates by using the copular \texttt{IS} construction, for instance \texttt{DRUNK} in the example. We deal with reflexive pronouns by introducing the appropriate pronominal link, and just restating the noun, for instance \texttt{HIMSELF} in the example.  We obtain a text diagram for each slice with gates:
\[
\scalebox{0.8}{\input{circ2textex2.tikz}}
\]
Then we join the diagrams of each slice with pronominal links that mirror the connectivity of the wire twisting slices:
\[
\scalebox{0.8}{\input{textualisation3.tikz}}
\]
The structure of pronominal links in text coincides with the compositional structure of circuits. By the rules of Section \ref{sec:congraphas} relating text diagrams to hybrid grammar, and by the interpretation rules of Refinement \ref{asm:conj}, we obtain a hybrid grammar text:
\[
\scalebox{0.8}{\input{circ2textexFIN.tikz}}
\]
\[
\footnotesize{
\texttt{\underline{ALICE SEES THAT BOB IS DRUNK AND ALSO THAT CLAIRE EXISTS.}}
}\]
\[
\footnotesize{
\texttt{\underline{ALICE TELLS CLAIRE THAT DENNIS HATES DEE AND ALSO THAT DEE LIKES DENNIS, BOB LAUGHS AT BOB.}}
}\]
By construction, our rewrite rules will turn the text back into our starting circuit.
\end{proof}
\end{proposition}

\begin{example}
In Figure \ref{fig:comic3} we give the example of textualisation of a text circuit that appeared in all of our previous such illustrations.

\begin{figure}[h!]
    \centering
    \scalebox{0.75}{\input{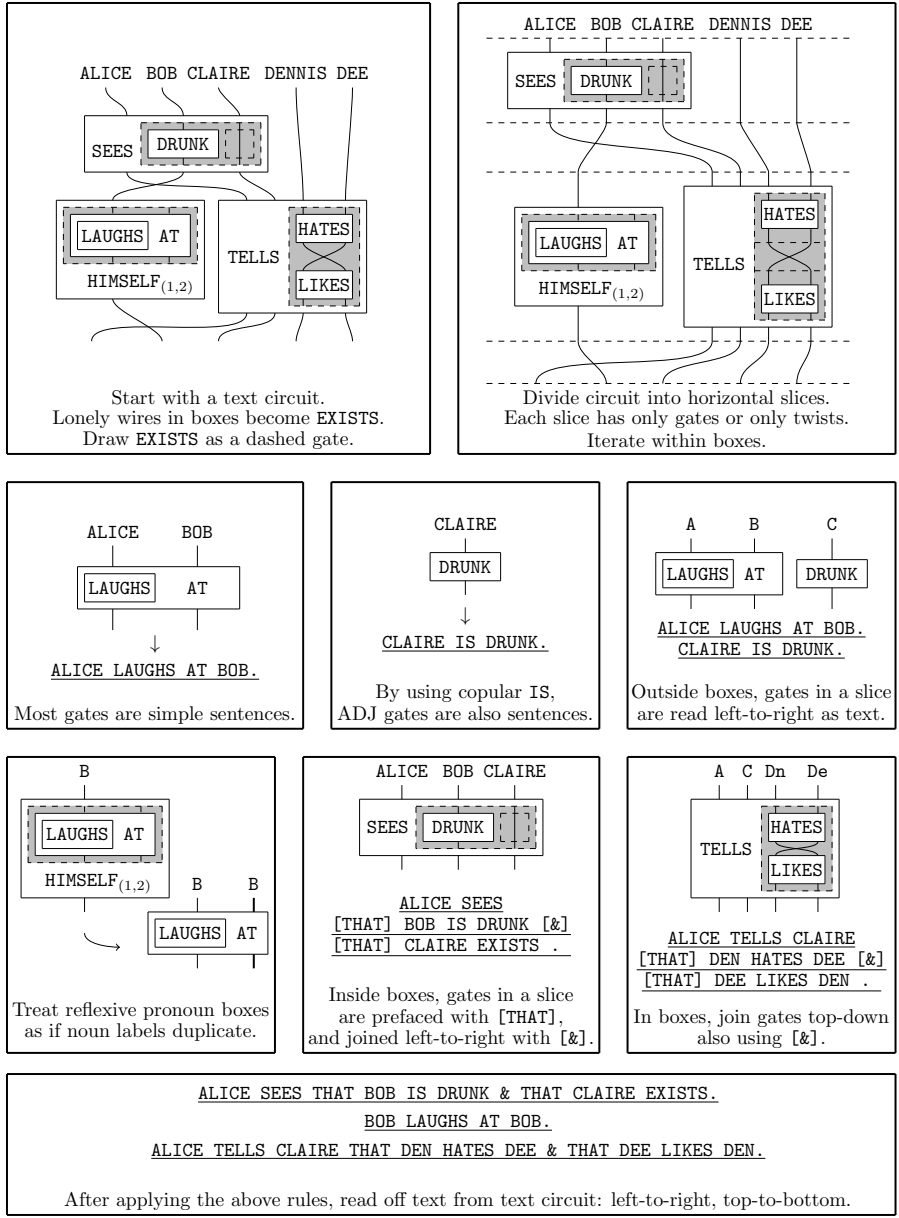}}
    \caption{Textualisation of circuits, illustrated.}
    \label{fig:comic3}
\end{figure}
\end{example}

\subsection{Extending grammar by means of equations}

An immediate consequence of the Text Circuit Theorem is the existence of a nontrivial equivalence relation on parsed text. In prose: we consider two texts equivalent if they result in equal text circuits.  So we may define  equivalence of texts $\mathcal{T}_1 \equiv \mathcal{T}_2$ as follows:
\[
\left.\begin{array}{c}
\mbox{exists rewrite}\ \  \mathcal{T}_1 \Rightarrow \mathcal{C}_1\\ 
\mbox{exists rewrite}\ \  \mathcal{T}_2 \Rightarrow \mathcal{C}_2 
\end{array}\right\} \ \ \mbox{such that}\ \  \mathcal{C}_1 =_{\mathfrak{C}} \mathcal{C}_2  
\] 
where $\mathcal{T} \Rightarrow \mathcal{C}$ denotes a rewrite of text with grammar $\mathcal{T}$ to text circuit $\mathcal{C}$, and $\mathcal{C}_1 =_{\mathfrak{C}} \mathcal{C}_2$ denotes equality of text circuits -- cf.~Convention \ref{conv:sliding}.

We have seen so far how text circuits appear to capture the essential connectivity of meaning underneath the bureaucracy of natural language. By this consideration, the equivalence relation on text we have defined looks a lot like `meaning equivalence'. Now we turn this observation into a guiding assumption.

\begin{thesis}[Text Circuit Thesis]
Equal circuits stand for equal text meanings. 
\end{thesis}
 
The thesis provides a recipe for engineering extensions of our grammar  
to accommodate grammatical phenomena beyond what we have considered here. We will explicitly demonstrate the inclusion of passive voice, possessive pronouns, and adjectivalisation of verbs using \texttt{[ING]} below. 

When it is broadly agreed that $\mathcal{T}_1$ and $\mathcal{T}_2$ `mean the same thing',  then we will \underline{postulate} that to be the case, and from this reverse engineer what the pieces of the text diagrams and corresponding rewrite rules of the grammar should be.  If $\mathcal{T}_1$ makes use of a grammatical phenomenon $\mathcal{GP}$ beyond what we have considered thus far, while $\mathcal{T}_2$ is within our diagrams/grammar, then we make the circuits of $\mathcal{T}_1$ and $\mathcal{T}_2$ equal by adding  the pieces of the text diagrams and corresponding rewrite rules for $\mathcal{GP}$. This procedure is best illustrated by the following examples, the first three of which are explored in \cite{GramEqs}.\footnote{In \cite{GramEqs} rather than extending diagrams/grammar by postulating equations like we do here, we used grammatical structure and so-called `internal wirings' \cite{FrobMeanI, FrobMeanII, GrefSadr, KartsaklisSadrzadeh2014, CLM, CoeckeText, CoeckeMeich} to derive equations between sentences.  This was in fact the path we took in order to arrive at the results that we present here, and therefore the amount of credit we give here to those internal wirings.  A forthcoming paper will be dedicated to this broad topic of `internal wirings' and their relationship to the results presented here \cite{IntWire}, and we also briefly address them in Section \ref{sec:discussion}.}

\begin{example}[Passive voice]
In English, expressing a transitive verb in passive voice reverses the order of subject and object. For example, the following sentence in passive voice:
\[
\texttt{\underline{BOB \bB IS LIKED BY\e ALICE}}
\]
conveys the same factual data as the following sentence in active voice:
\[
\texttt{\underline{ALICE \bB LIKES\e BOB}}
\]
To accommodate the passive voice, we introduce a new wire type $\texttt{TVP}_{psv.}$. In the figure below, on the left we introduce a string diagram that allows a \texttt{TVP} to become a $\texttt{TVP}_{psv.}$ within a phrase scope, and on the right, we introduce a string diagram that closes the passive voice phrase scope by labelling with \texttt{\underline{IS}} and \texttt{\underline{BY}}:
\[
\input{passivevoicegen1.tikz}
\]
We also ask that every label: 
\[
\begin{tikzpicture}
	\begin{pgfonlayer}{nodelayer}
		\node [style=none] (37) at (0, 0.5) {};
		\node [style=none] (38) at (0, -0.5) {};
		\node [style=none] (40) at (0, -1) {\texttt{\underline{TV}}};
		\node [style=none] (43) at (0, 1) {\texttt{TVP}};
	\end{pgfonlayer}
	\begin{pgfonlayer}{edgelayer}
		\draw [style=KL] (37.center) to (38.center);
	\end{pgfonlayer}
\end{tikzpicture}

\]
has a passive voice counterpart:    
\[
\input{passivevoicegen3.tikz}
\]
e.g.~\texttt{\underline{LIKES}} becomes \texttt{\underline{LIKED}}.
We also stipulate that, apart from the diagram on the left above, $\texttt{TVP}_{psv.}$ mimics \texttt{TVP} for all other diagrams. This is so that we may further generate text grammar/diagrams  
within the passive voice scope to obtain, e.g.:
\[
\texttt{\underline{BOB IS DEEPLY LIKED BY ALICE}}
\]
As a point of justification, the exception for the left diagram and having a separate $\texttt{TVP}_{psv.}$ is necessary to disallow the passive construction to be applied repeatedly, which for example overgenerates the \bR\textbf{ungrammatical}\e:
\[
\bR \texttt{\underline{BOB IS IS LIKED BY BY ALICE}} \e
\]
Returning to the example, we reformulate instances of passive voice into active voice by the following rewrite rule:
\[
\input{passiveredux.tikz}
\]
So we obtain the following proof of text equivalence:
\[
\input{passivecircexcopy.tikz}
\]
\end{example}  

\begin{example}[Possessive Pronouns]\label{ex:posspron}
Consider the possessive pronoun \texttt{HIS} in the phrase:
\[
\texttt{\underline{BOB DRINKS HIS BEER}}
\]
We can reformulate the noun phrase \texttt{HIS BEER} as the noun phrase \texttt{BEER THAT HE OWNS} without changing the core meaning. Applying this transformation, we obtain:
\[
\texttt{\underline{BOB DRINKS BEER THAT HE OWNS}}
\]
We know that the relative pronoun \texttt{THAT} in this case gives us the same circuit as the text:
\[
\texttt{\underline{BOB DRINKS BEER. BOB OWNS BEER.}}
\]
We can make this text equivalent to the original \texttt{\underline{BOB DRINKS HIS BEER}} by introducing the following generation and rewrite rules for possessive pronoun labels. We introduce a dotted-blue possessive pronominal link type with two generators, so that the possessor is pronominally referred to by the possessive pronoun generator:
\[
\input{possprongen.tikz}
\]
The rewrite rule that eliminates the possessive pronoun and the possessive pronominal is as follows:
\[
\input{posspronredux.tikz}
\]
So, returning to our sentence, we have a series of rewrites from text diagram to text circuit: 
\[
\input{posspronexample.tikz}
\]
Note that while text diagrams from text are planar, in the process of rewrites, we may freely twist wires.\end{example}

\begin{example}[Adjectivalisation of Verbs by gerund \texttt{-ING}]
Appending \texttt{-ING} to a verb allows that verb to be used as if it were an adjective. For example, the noun phrase:
\[
\texttt{\underline{DANC-ING ALICE}}
\]
may be reformulated as the noun phrase: 
\[
\texttt{\underline{ALICE WHO DANCES}}
\]
We model \texttt{[ING]} as a generator that transforms an \texttt{ADJ} into a \texttt{IVP}.
\[
\input{inggen.tikz}
\]
The reduction rule is straightforward:
\[
\input{ingredux.tikz}
\]
As a result, we obtain equivalences between the sentences:
\begin{itemize}
    \item \texttt{\underline{ALICE IS DANCING.}}
    \item \texttt{\underline{ALICE DANCES.}}
\end{itemize}
and the noun-phrases:
\begin{itemize}
    \item \texttt{\underline{DANCING ALICE}}
    \item \texttt{\underline{ALICE WHO DANCES}}
\end{itemize}
\[
\input{ingexamples.tikz}
\]
\end{example}

\section{Discussion}\label{sec:discussion}

\subsection{Parsing text}\label{sec:parsing}

As we mentioned in the introduction, we also want machines to be able to distil circuits from real text. Therefore in general distillation would rather look as follows:
\[
\input{Distil2.tikz}
\]
where parsing is the process of automatically assigning grammatical and other structure to sentences and text.  As described in \cite{liu_language_2021}, common dependency parsers such as SpaCy \cite{spacy2} behave well enough on simple sentences --  handling the phrase structure and phrase scope components -- to give output that can be interpreted systematically as individual gates that constitute a text circuit.
For our purposes, the additional  
requirement for parsing real text is a procedure for identifying pronominal links -- a problem known as \emph{anaphora resolution}, for which algorithms have existed for some time \cite{lappin-leass-1994-algorithm}. In fact, modern ML language models are approaching human-level performance for this task \cite{brown_language_2020}.  Sentences with relative pronouns can be systematically decomposed into multiple simple sentences with pronominal links by pattern-matching. This obtains a list of simple sentences, pronominally linked.

\subsection{Arbitrary text}

Recall that we have only dealt with a fragment of language here, but one that we have demonstrated can be extended to accommodate more aspects of language via implicit equations. The process of parsing arbitrary text may require text to be preprocessed against these implicit equations to obtain a list of simple sentences. To qualitatively extend our fragment beyond what can be done with implicit equations, in future work we will also incorporate grammar that requires higher order representations besides operations that act on gates, to capture for instance quantifiers and determiners.

%

\subsection{Relationships to other syntactic and semantic formalisms}

In this section, we begin to compare and contrast our hybrid grammar, text diagrams, and text circuits to some leading formalisms in the literature. We devote the bulk of analysis to text circuits. This section also helps to illustrate that while we obtain text circuits by starting from hybrid grammar, one could also use other established formalisms (dependency grammar, typelogical grammar, discourse representation theory etc.) as a starting point for obtaining text circuits.
Indeed, for practical applications where one wants to generate text circuits, this is likely to be the easiest path.

\subsubsection{Our hybrid grammar for text}\label{subsec:hybgram}

The main ingredient in the passage from single sentences to text is the pronominal link data, which we assume is obtained by separate means -- Section \ref{sec:parsing}. More research is needed to formally relate hybrid grammar to extant formalisms in the literature, but here we suggest that hybrid grammar belongs in the class of grammars subsumed by linear context-free rewrite systems \cite{weir_characterizing_1988}, which contains for instance linear indexed grammar \cite{vijayashanker_study_1987}, combinatory categorial grammar \cite{weir_combinatory_1988}, and head grammar \cite{weir_relationship_2002}.

We have two main reasons to suspect this. First, that the so-called composition functions of linear context-free rewrite systems closely resemble the gates of text circuits, as both are linear and regular in their input arguments.

Second, if we suppose that the algorithm for determining pronominal links is `effectively computable' in the sense identified with `polynomial time complexity' (known as Cobham's thesis), then parsing text in hybrid grammar is evidently polynomial time. One characterisation of mild-context sensitivity in \cite{zwicky_tree_1985} is just polynomial time parsing, and the development of linear context-free rewrite systems is also motivated by a sharper characterisation of mild-context sensitivity in these terms.

We leave further, formal exploration of these correspondences for future work.  

\subsubsection{Text diagrams}\label{subsec:textdiag}

\paragraph{Relation to dependency grammars.} Text diagrams geometrically and conceptually resemble dependency grammars \cite{tesniere_elements_2015}, and more generally link grammars \cite{sleator_parsing_1995}. This relation is elaborated upon in \cite{liu_language_2021}, and used as a theoretical basis for a `handwritten' text-to-circuit parsing algorithm.

Text diagrams as we have used them in this paper are much lighter than dependency grammars in that they do not specify as many grammatical relationships. Recalling the discussion of `two-dimensionality' in the introduction, this is partly due to the fact that the geometry of diagrams `absorbs' some of the grammatical roles such as subjects, objects, and referents of relative pronouns.
Importantly, this minimalism does not come at the cost of asceticism. As we have seen in Example \ref{ex:posspron}, one may add new wire types and rewrite rules to capture grammatical relationships as desired.

\paragraph{Relation to typelogical grammars.}  In work under preparation we will show how text diagrams for individual sentences may arise from pregroup grammars \cite{LambekBook}, which are known to be equivalent (cf.~above) to context-free grammars \cite{buszkowski_pregroup_2008}, and intimately related to combinatory categorial grammars in application \cite{DRichie}.

\paragraph{Relation to transformational grammars.} The rewrite rules of text diagrams are a type-restricted analogue of transformational grammars. Unrestricted transformational grammars are known to be turing complete \cite{peters_generative_1973} and hence too powerful to characterise natural language. We leave a formal characterisation of text diagrams and their placement on the Chomsky hierarchy for future work.

\subsubsection{Text circuits.}  We sketch arguments that relate text circuits to different approaches to theoretical and computational linguistics, prefacing each paragraph with the nature of the relation.

\paragraph{Refinement of discourse-representation.}
The approach to modelling the meaning of text (or \textit{discourse}) that is embodied by text circuits closely aligns with the dynamic semantics perspective.
The shared philosophy is that pieces of text are viewed as updating an existing context with new information, the result of which is an updated context.
Among the theories of dynamic semantics, text circuits bear a particularly close resemblance to discourse representation theory \cite{kamp_discourse_2010}.
Roughly speaking, the \texttt{NP} wires of text circuits correspond to the discourse referents of DRT, and the gates which act upon the \texttt{NP} wires correspond to the predicates and other conditions upon the referents that appear in DRT.
It is worth noting that the notion of discourse referent goes all the way back to the classic paper \cite{karttunen1976discourse}, where it was introduced in the context of a hypothetical algorithm which parses a text and accordingly records all the information learnt about each referent.
In the NLP context, text circuits can be seen as offering a direct realisation of this aim.

As a key point of difference, discourse-level theories typically take the syntax and semantics of sentences to be a given, and from there establish theories of grammatical text. In this sense, text circuits are an extension of discourse theory down to accommodate sentence grammar.
Accordingly, one domain in which text circuits surpass discourse representation theory is in the handling of adverbs as boxes, which are naturally interpreted as higher-order operators. In contrast, we are not aware of a variant of discourse representation theory where discourse boxes are decorated with the data of higher-order predicates, nor are we aware of a standard approach to higher-order predication in logic.

\begin{figure}[h!]  
    \centering
    \input{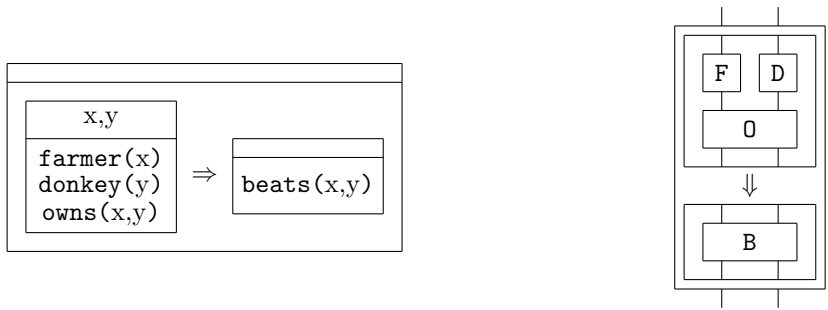}  
    \caption{Left: a discourse representation structure for the sentence ``If a farmer owns a donkey he beats it". Right: a text circuit for the same sentence.}
    \label{fig:donkeysentence}
\end{figure}

We demonstrate with a simple example the correspondence that allows text circuits to be viewed as a `fine-graining' of the discourse representation structures of DRT. Simple sentences are depicted as (possibly nested) gates in text circuits, essentially in correspondence with complex conditions of DRT represented by nested DRS boxes. The distinction is that while box-nesting is the only form of composition in DRT, text circuits also have access to the diagrammatic composition of monoidal categories, which graphically elaborates the content of discourse boxes according to text progression, illustrated in Figure \ref{fig:donkeysentence}.  

Given the close analogy between DRT and DisCoCirc, we believe further exploration of the relationship between the two will be fruitful. In particular, we believe that such a study will help expand the capabilities of text circuits to model a larger fragment of language and deal with more complicated linguistic phenomena. This assumption comes from the fact that many of the challenges in modelling text with text circuits are exactly challenges that were faced by DRT.

\begin{figure}[h!]    
    \centering
    \input{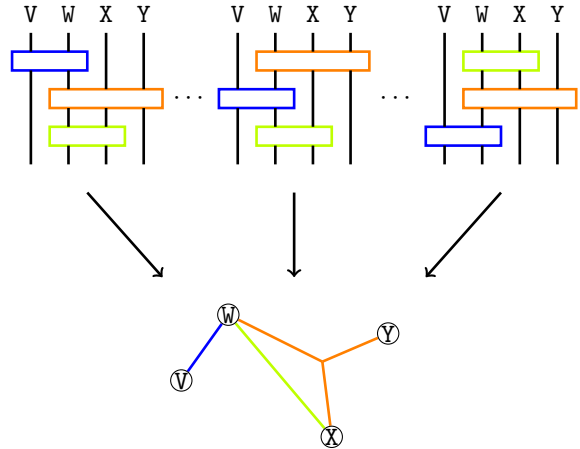}
    \caption{Top: text circuits with the same gates, but in different orders. Bottom: a graph representation obtained by stipulating that gate order does not matter.}
    \label{fig:decommutate}
\end{figure}

\paragraph{Refinement of graph-representation semantics.}
Graphs are a common computer implementation of knowledge representation. Text circuits can be viewed as graph models with the refinement of noncommutativity that reflects the order of text. That is to say, one can always interpret the gates of a text circuit to commute, which obtains the equivalent of a graph model, illustrated in Figure \ref{fig:decommutate}.

There are two benefits to noncommutativity: one theoretical, and one practical. The theoretical benefit, as discussed above, is that one can always forget the order of facts in the text to obtain graph representations, but not vice versa: `flat' graph representations of text meaning reflect how in classical logic, for perfect logicians, the order of facts does not affect the outcome of inference. The practical benefit regards this last point; we know that there are order-effects in human reasoners, just as we know that jokes do not work when the punchline comes first. Noncommutativity is practically essential to capture the \emph{process} of reading and understanding text, rather than merely outcomes.

\paragraph{Cousin of Montague grammar.}
Montague semantics \cite{montague1970universal} is (glossed) an \emph{algebra homomorphism} from syntax to semantics \cite{janssen_montague_2021}. Algebra refers to the mathematical framework for composition, and homomorphism to a mapping that preserves the composition structure of that algebra. The difference between text circuits and Montague semantics arises from different notions of what the algebra of composition ought to be. Montague's algebras were \emph{clones}, which are equivalent \cite{kerkhoff_short_2014} to Lawvere theories \cite{andre_algebraic_1968}, which are in turn special cases of PROPs \cite{noauthor_prop_nodate}. The fundamental distinction between Lawvere theories and PROPs is that the former is many-input-single-output, and the latter is many-input-many-output. Text circuits are fully compatible with Montague's fundamental idea of a `structure-preserving map' from syntax to semantics. In fact, the success of recent quantum natural language processing experiments \cite{QNLPPlus100} may be attributed to just such a `structure-preserving' implementation of text circuits on a quantum computer.

\section{Acknowledgements}

We thank Dimitri Kartsaklis, Konstantinos Meichanetzidis, Anna Pearson, Robin Lorentz, Alexis Toumi, Giovanni de Felice, Ian Fan, Richie Yeung, Stephen Clark and the SEMSPACE referees for useful feed-back on a (very different) previous version of this paper. We thank Vid Kocijan, Richie Yeung, Amar Hadzihasanovic, Stephen Clark and Michael Moortgat for comments on the current version.

\bibliographystyle{plain}
\bibliography{mainNOW}

\end{document}